%% file: acl_latex.tex
\pdfoutput=1
\documentclass{article}
\usepackage{CJKutf8}

\usepackage{acl}
\usepackage{booktabs}
\usepackage{graphicx}
\usepackage{enumitem}
\usepackage{wrapfig}
\usepackage{algorithm}
\usepackage{algpseudocode}
\usepackage{amssymb}

\usepackage{microtype}
\usepackage{amsmath}
\usepackage{colortbl}
\usepackage[most]{tcolorbox}
\usepackage[utf8]{inputenc}
\definecolor{lightgray}{rgb}{0.9,0.9,0.9}
\usepackage{caption}
\usepackage{subcaption}
\usepackage{xcolor}
\usepackage{setspace}
\usepackage{url}
\usepackage{multirow}
\usepackage{colortbl}
\usepackage{tabularx}
\usepackage{blindtext}
\usepackage{pgfplots}
\pgfplotsset{compat=1.18} 
\usepackage{tikz}
\usetikzlibrary{er,positioning,bayesnet}
\usepackage{makecell}
\usepackage{tipa}
\usepackage{siunitx}
\usepackage{nicefrac}
\usepackage{tocloft}
\usepackage{listings}
\usepackage{xltabular}
\usepackage{adjustbox}
\usepackage{xurl}
\usepackage{times}
\usepackage{latexsym}
\usepackage{twemojis}
\usepackage{balance}
\usepackage[T1]{fontenc}
\usepackage[utf8]{inputenc}

\usepackage{microtype}

\usepackage{inconsolata}

\usepackage{graphicx}

\title{RedOne: Revealing Domain-specific LLM Post-Training\\ in Social Networking Services}

\author{
Fei Zhao, Chonggang Lu, Yue Wang, Zheyong Xie, Ziyan Liu, Haofu Qian,\\
\textbf{ JianZhao Huang, Fangcheng Shi, Zijie Meng, Hongcheng Guo, Mingqian He, Xinze Lyu, }\\
\textbf{Yiming Lu, Ziyang Xiang, Zheyu Ye,  Chengqiang Lu,  Zhe Xu,  Yi Wu, Yao Hu,}\\
\textbf{  Yan Gao, Jun Fan,  Xiaolong Jiang, Weiting Liu, Boyang Wang, Shaosheng Cao\thanks{Corresponding author.}} \\
NLP Team, Xiaohongshu Inc.,  China \\
\texttt{caoshaosheng@xiaohongshu.com}
}

\begin{document}
\maketitle
\begin{abstract}

As a primary medium for modern information dissemination, social networking services (SNS) have experienced rapid growth, which has proposed significant challenges for platform content management and interaction quality improvement. Recently, the development of large language models (LLMs) has offered potential solutions but existing studies focus on isolated tasks, which not only encounter diminishing benefit from the data scaling within individual scenarios but also fail to flexibly adapt to diverse real-world context. To address these challenges, we introduce \textbf{RedOne}, a domain-specific LLM designed to break the performance bottleneck of single-task baselines and establish a comprehensive foundation for the SNS. RedOne was developed through a three-stage training strategy consisting of continue pretraining, supervised fine-tuning, and preference optimization, using a large-scale real-world dataset. Through extensive experiments, RedOne maintains strong general capabilities, and achieves an average improvement up to 14.02\% across 8 major SNS tasks and 7.56\% in SNS bilingual evaluation benchmark, compared with base models. Furthermore, through online testing, RedOne reduced the exposure rate in harmful content detection by 11.23\% and improved the click page rate in post-view search by 14.95\% compared with single-tasks finetuned baseline models. These results establish RedOne as a robust domain-specific LLM for SNS, demonstrating excellent generalization across various tasks and promising applicability in real-world scenarios.
\end{abstract}

\input{section/intro}

\input{section/related_work}
\input{section/continue_pretrain}

\input{section/alignment}
\input{section/experiment}
\input{section/conclusion}

\bibliography{custom}

\newpage
\appendix

\input{section/appendix}

\end{document}

%% file: section/intro.tex
\section{Introduction}

With the widespread adoption of online platforms and mobile applications, social networking services (SNS) have emerged as a central medium for modern information dissemination, such as communication, knowledge sharing, and emotional expression~\cite{elahimanesh2025emotion}. Unlike the general textual corpora, SNS data is highly informal, context-sensitive, and often emotionally charged. These characteristics present unique challenges including linguistic variability, frequent role-switching, and subtle conversational norms, which complicate applications (e.g. platform content management and interaction quality improvement) for traditional natural language processing (NLP) systems~\cite{jin2024mm}.

\begin{figure}[tbp]
    \centering
    \includegraphics[clip, width=0.48\textwidth]{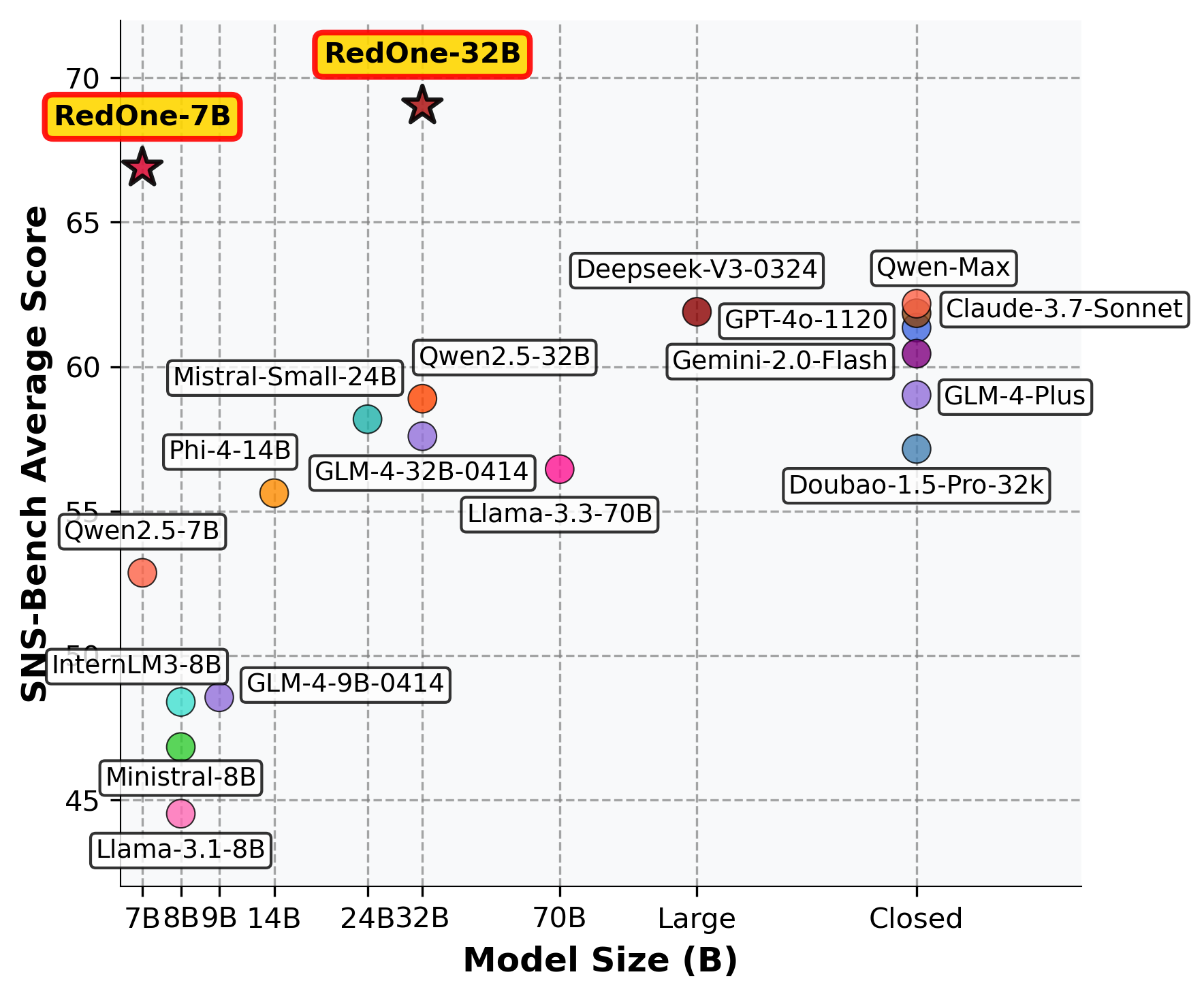}
    \caption{Performance comparison of different models in the SNS domain, where all models are instruction-tuned and the evaluation score is the average of all tasks on SNS-Bench.}
    \label{fig:intro}
\end{figure}

Given these complexities, numerous studies have explored recent advanced large language models (LLMs) based adaptation for SNS-related tasks~\cite{zeng2024large, jiang2023social}. However, these solutions primarily focus on isolated tasks, which not only experience diminishing benefits as data scales within individual scenarios but also struggle to adapt flexibly to diverse real-world contexts. This highlights a fundamental limitation in current SNS domain-specific models, where performance plateaus due to the inability to incorporate a more diverse domain knowledge corpus during training \cite{yue2025does}.

To address these deficiencies, we introduce \textbf{RedOne}, a demain-specific LLM with a meticulous three-stage post-training strategy using a large-scale dataset from real-world, which consists of continued pretraining (CPT), supervised fine-tuning (SFT), and preference optimization (PO). In the CPT stage, the model acquire extensive foundational knowledge in the SNS domain by processing large-scale corpora. Building on this foundation, the SFT stage refines the model's capability to tackle specific SNS tasks by leveraging carefully defined domain-specific problem formulations. Finally, in the PO stage, we further optimize the model's behavior to ensure seamless alignment with human preferences and maximize its practical utility in real-wold deployments.

Through extensive experiments, RedOne not only maintains strong general capabilities, but also excels across multiple SNS-specific evaluation benchmarks, significantly outperforming leading proprietary or open-source models as shown in Figure \ref{fig:intro}. Further online testing in harmful content detection and post-view search, indicates its broad and promising potential application in real-world scenarios.

Our contributions can be summarized as follows:

\begin{itemize}
    \item We introduce RedOne, a domain-specific LLM, engineered to break the performance bottleneck of single-task models, providing comprehensive improvements for SNS. 
    \item A three-stage training strategy is designed, using a large-scale real-world dataset, which maintains strong general capabilities while delivering exceptional generalization across diverse SNS tasks. 
    \item Through extensive experiments and online testing to demonstrate RedOne's effectiveness across a wide range of tasks, and establish a comprehensive and robust baseline for SNS application.
\end{itemize}

%% file: section/related_work.tex
\section{Related Work}

\subsection{NLP tasks in Social Networking Services}    
Due to the inherent characteristics of SNS platforms, namely their informality and rapid linguistic evolution~\cite{carr2015social}, these platforms present numerous complex NLP challenges that have garnered sustained academic attention. In the early stages of development, researchers primarily focused on fundamental capability assessments, particularly prevalent tasks such as sentiment analysis~\cite{mohammad2018semeval, rosenthal2019semeval}, harmful content detection~\cite{i2019multilingual, lu2024towards}, and meme detection~\cite{10.1145/3612920, lin2024towards}. Following the emergence of LLMs and building upon previous research foundations, various techniques have evolved in multiple domains, including content understanding~\cite{kumar2024watch,kmainasi2024llamalens}, information extraction~\cite{islam-goldwasser-2025-uncovering,li2024socialgpt,peng2024metaie}, and dialogue systems~\cite{yi2024survey,zhang-etal-2024-self-emotion}. These technological advances have significantly enhanced problem-solving capabilities within the SNS domain, but have primarily focused on single tasks. In contrast to these works, RedOne demonstrates superior performance across diverse SNS tasks, providing a foundational model for improved services.

\subsection{Domain-specific Post-training}

To better serve specialized domains, recent efforts have focused on developing vertical domain LLMs across various fields, including finance~\cite{wu2023bloomberggpt, konstantinidis2024finllama}, law~\cite{colombo2024saullm}, home renovation~\cite{wen2023chathome}, medicine~\cite{xiong2023doctorglm, chen2023bianque, yang2024zhongjing, wu2024pmc, zakka2024almanac}, and scientific research~\cite{azerbayev2023llemma, bi2023oceangpt, yang2024pllama}. Despite these advancements, these vertical domain LLMs have not addressed the unique challenges posed by SNS. While ~\cite{liu2024can} and~\cite{yang2024mentallama} explore the application of LLMs to a limited set of NLP tasks within SNS, their coverage remains constrained. Therefore, a significant gap exists in this area, which RedOne aims to address.

%% file: section/continue_pretrain.tex
\section{RedOne Model}

\begin{figure*}
    
    \centering
    \includegraphics[width=0.88\linewidth]{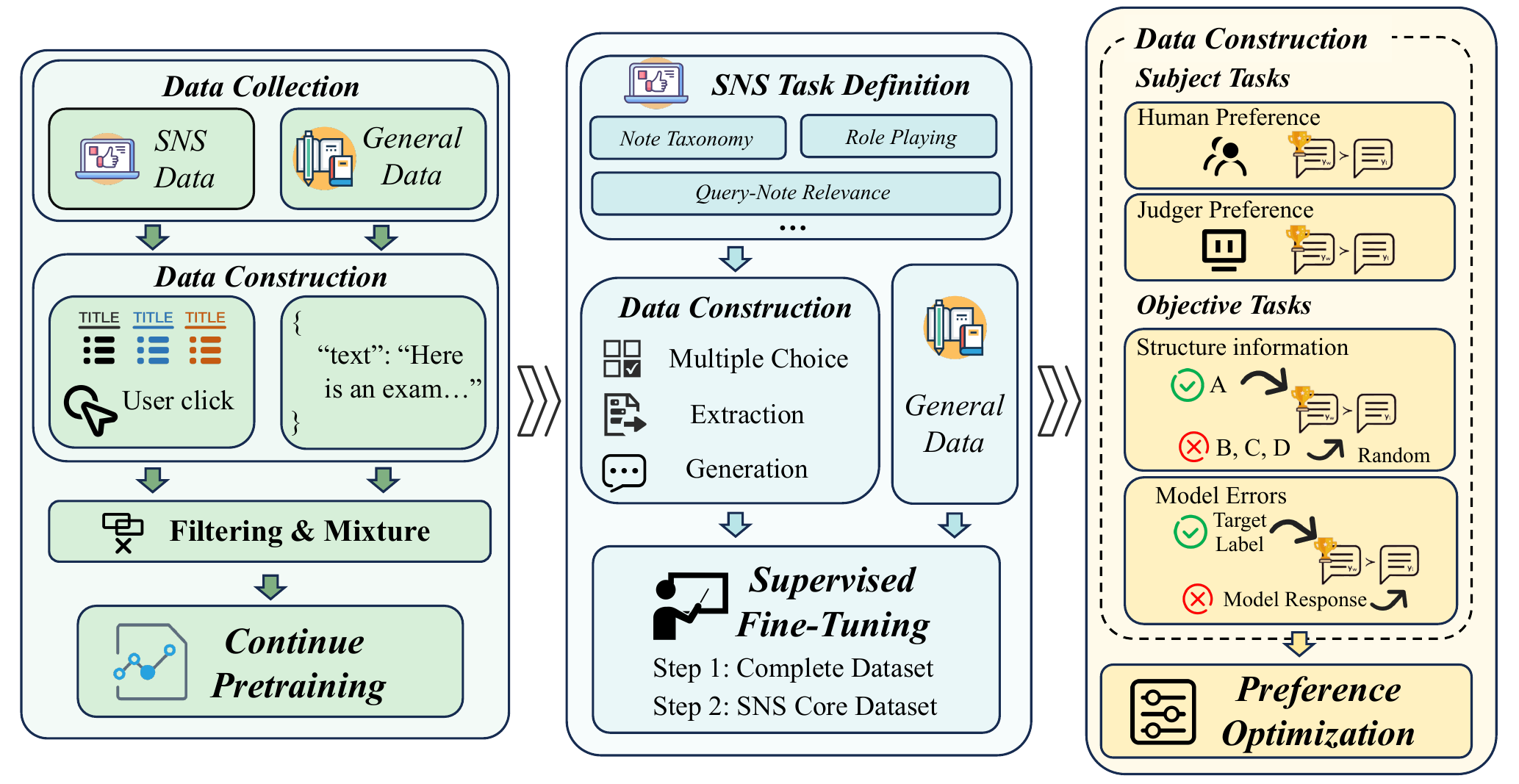}
    \caption{Overview of our training pipeline.}
    \label{fig:pipeline}
\end{figure*}

As illustrated in Figure \ref{fig:pipeline}, the training strategy of RedOne contains three stages. First, in Section 3.1, we conduct continue pretraining to enrich the model's grasp of nuanced SNS field knowledge. Subsequently, in Section 3.2, we sharpen the model's instruction-following capabilities through supervised fine-tuning across various tasks. Finally, we leverage preference information from the training data to perform preference optimization, ultimately yielding the RedOne with superior performance in the SNS domain.

\subsection{Continue Pretraining}

To enhance the large model's fundamental domain knowledge 
we conducted continue pretraining at this stage, 
which can be divided into three sub-stages: data collection and data construction, filtering and mixture, along with domain-aware continue pretraining.

\label{sec:pre}
\subsubsection{Data Collection and Data Construction}

We specifically collected continue pretraining data from the following two data sources: 

(1) \textbf{General High-quality Data}. We selected several high-quality open-source pretraining corpora~\cite{qiu2024wanjuan,weber2024redpajama,penedo2024fineweb} to preserve the model's fundamental generalization capabilities. 
To improve training efficiency, we uniformly construct all general data into single-sentence text format and perform segmentation and concatenation processing based on predefined text length thresholds.

(2) \textbf{SNS-specific Domain Data}. We collect the large-scale training data from SNS platforms and the open web, capturing diverse social communication patterns including informal discussions, short-form comments, sarcasm, emotionally charged content, and so on. For better reveal the underlying information in the pretraining data, we incorporate user interaction data to guide the training process. Specifically, we group contexts and comments with their corresponding user interaction data, which naturally clusters semantically related SNS content without additional processing. 
Through these steps, we collected and constructed a large-scale dataset comprising various tasks with over 100B tokens for downstream processing.

\subsubsection{Filtering and Mixture}
Considering data quality is crucial for model training~\cite{zhou2023lima}, we constructed a data-filtering pipeline inspired by \cite{yuan2024continued}, which comprises task-oriented rule filtering and small language model filtering~\cite{wang2025ultra}. The former identifies specific error content such as HTML tags and repetitive sentences, while the latter focuses on global assessment aspects including coherence and tone appropriateness. Based on this data-filtering pipeline, we further applied the RegMix method ~\cite{liu2024regmix} to identify an optimal data mixture distribution and filter out unnecessary data.
Through this comprehensive filtering and mixture process, we ultimately constructed a high-quality dataset of 20B tokens for training.

\subsubsection{Domain-aware Continue Pretraining}

After data construction, we conduct continue pretraining on the complete dataset. Specifically, RedOne is trained from the Qwen2.5~\cite{qwen2025qwen25technicalreport} checkpoint using the same configurations, leveraging its strong linguistic capabilities across multiple domains. Through this domain-aware continue pretraining process, we ultimately obtain a model that effectively captures SNS-specific linguistic patterns while maintaining minimal degradation in general language modeling capabilities.

%% file: section/alignment.tex
\subsection{Supervised Fine-Tuning}

    To bridge the gap between pretraining objectives and the specific requirements of real-world SNS applications, we further conduct supervised fine-tuning on our model through carefully designed data construction and multi-stage training strategies.

\subsubsection{Task Definition and Data Construction}

As SFT training data is significat affect the final instruction following ability in domain tasks~\cite{dong2023abilities}, we extensively collected large-scale user-generated content from public platforms, including notes, comments, queries, and interaction logs, which provide real enviorment signal for us to improve model actions. Notably, we focused on preserving the linguistic style which exhibit typical SNS characteristics such as informal language, sarcasm, sentiment, and topical shifts while collecting data~\cite{eisenstein2013bad}, aim for representative and practical coverage for SNS scenarios. 

After data collection, we ultimatly consolidate six kinds of core capabilities essential for SNS applications: content understanding, information extraction, semantic matching, user behavior modeling, dialogue and persona simulation, and translation, as show in Table~\ref{tab:sns_tasks_simple}. Each is supported by well-defined tasks reflecting real-world challenges and the overview is shown in Appendix~\ref{appendix:sft_data_statistics}.

Additionally, during SFT, we also incorporated open source instruction data covering general tasks such as instruction following~\cite{li2025infinity,zhou2023lima}, multiturn dialogue~\cite{zhao2024wildchat}, and long chain-of-thought (CoT) reasoning~\cite{guha2025openthoughts,ye2025limo}, to mitigate catastrophic forgetting ~\cite{mccloskey1989catastrophic} and retain generalization ability of RedOne model. 
% Training Prompt templates of all tasks can be found in Appendix~\ref{appendix:prompts}.

\begin{table}[t]
\centering
\small
\resizebox{0.48\textwidth}{!}{
\begin{tabular}{lc}
\hline
\textbf{Task Name} & \textbf{Capability} \\
\hline
Note Taxonomy & Content Understanding \\
Query Classification & Content Understanding \\
Query Intent Recognition & Content Understanding \\
Hashtag Prediction & Information Extraction \\
Machine Reading Comprehension & Information Extraction \\
Highlight Word Detection & Information Extraction \\
Query-Note Relevance & Semantic Matching \\
Query-Note Retrieval & Semantic Matching \\
Post-View Search & User Behavior Modeling \\
Emotional Companion Dialogue & Dialogue \\
Role-playing Dialogue & Dialogue \\
SNS Domain Translation & Translation \\
\hline
\end{tabular}
}
\caption{Overview of SNS Tasks and Their Capabilities}
\label{tab:sns_tasks_simple}
\end{table}

\subsubsection{Two-Step Training}

In domain SFT, a two-step mixed fine-tuning has been demonstrated to effectively enhance domain-specific capabilities~\cite{dong-etal-2024-abilities}. For RedOne's SFT, we implement this strategy by mixing SNS-specific data with general data across two steps.
In the first step, we train the model on the complete SNS dataset combined with a large volume of general data. This approach enables the model to learn diverse task formats within the SNS domain while preserving its generalization capabilities. In the second step, we fine-tune the model using a higher proportion of SNS domain data, thereby further enhancing performance on domain-critical tasks.

\subsection{Preference Optimization}

SNS tasks like query-note relevance modeling often produce multiple plausible but quality-diverse outputs. While SFT improves instruction-following, it fails to exploit implicit preference signals among these candidates, causing overfitting and poor generalization~\cite{chu2025sft}.
To address these limitations, in this section, we carefully craft preference data and perform PO to obtain a better domain-specific model.

\subsubsection{Preference Data Construction}

To enhance alignment with human preferences and utilize the information embedded in data labels, we integrate different preference pair  construction strategies according to the nature of different task types. Specifically, we categorize our data into two types and adopt corresponding strategies:

For subjective tasks, such as emotional dialogue and role-playing, our primary objective is to achieve better alignment with human preferences. Therefore, the first step begins with domain experts creating preference annotations on model-generated responses~\cite{ouyang2022training}. Furthermore, to scale up the preference dataset, we evaluate the consistency between trained judge models~\cite{cao2024compassjudger} and human preference,  then leverage these models with high performance to expand specific data.

In contrast, for objective tasks with definitive correct answers, our strategy shifts toward extracting and utilizing the implicit structural information within the data labels. Here, we employ two approaches: First, we leverage the inherent structure of questions that contain both correct answers and incorrect options, constructing preference pairs that exploit the ordinal relationships within data. Complementarily, to actively address model limitations, we construct preference pairs from model errors, using ground truth as positive examples and incorrect predictions as negative to target specific weaknesses.

By integrating these tailored approaches, we systematically process all SNS-domain data according to their inherent characteristics, ultimately constructing preference optimization datasets that effectively capture both human preferences and implicit data information for comprehensive model enhancement.

\subsubsection{Direct Preference Optimization}

To effectively leverage the rich preference signals in our SNS dataset, we adopt DPO~\cite{rafailov2024direct} as our preference-based fine-tuning algorithm. This approach enables the model to better align with human preferences while simultaneously exploiting the latent information embedded in ground-truth labels. 

Finally, through this comprehensive three-stage training pipeline encompassing CPT, SFT and PO, we ultimately obtain a domain-specific large language model RedOne that demonstrates superior performance in the target domain while maintaining reasonable general capabilities.

%% file: section/experiment.tex
\section{Experiments}

\subsection{Implementation details}\label{sec:imp_details}
During the CPT stage, we follow the training process from Qwen2.5~\citep{qwen2} 
over a mixed corpus of general and SNS-specific data. SFT is conducted for three epochs in step one and two epochs in step two, with a maximum sequence length of \num{16384} using sequence packing, batch size of \num{128}, a linear warm-up ratio of 0.1. The learning rates are set according to model size: for the 7B model, we use \num{5e-6} in step one and \num{3e-6} in step two; for the 32B model, we use \num{3e-6} for both steps. Optimization is performed using AdamW~\cite{adamw} ($\beta_1{=}0.9$, $\beta_2{=}0.95$, $\epsilon{=}10^{-8}$). In the final PO stage, we employ a learning rate of $1\times10^{-7}$, batch size of \num{64}, sequence length of \num{4096}, training for two epochs, with SFT loss coefficient set to 0.3.

\input{section/7b_table}

\input{section/32b_table}

\subsection{Benchmarks}

For general capabilities evaluation, we use datasets similar to those employed in community, including general natural language comprehension (i.e. MMLU~\cite{mmlu}, CMMLU~\cite{cmmlu}, CEVAL~\cite{ceval}, GPQA-Diamond~\cite{gpqa}, NewsBench~\cite{newsbench}), reasoning (i.e. MMLU-Pro~\cite{mmlupro}, BBH~\cite{bbh}, GaokaoBench~\cite{gaokao-bench}), mathematics (i.e. AIME2025~\cite{aime25}, GSM8K~\cite{gsm8k} and MATH500~\cite{math}), coding (i.e. HumanEval~\cite{humaneval}, MBPP~\cite{mbpp}, and LiveCodeBench(2407\-2502)\cite{livecodebench}), translation (i.e. WMT-22/23/24 and Flores\cite{flores}), instruction following (i.e. IFEval~\cite{ifeval}), hallucination and human preference alignment (i.e. HaluEval~\cite{halueval} and CompassBench~\cite{compassbench}). To further evaluate RedOne's performance in SNS domain, we selected specialized SNS benchmarks including SNS-Bench~\cite{guosns} and SNS-TransBench~\cite{guo2025redefining}.

\subsection{Main Results}
\begin{figure}
    \centering
    \includegraphics[width=0.9\linewidth]{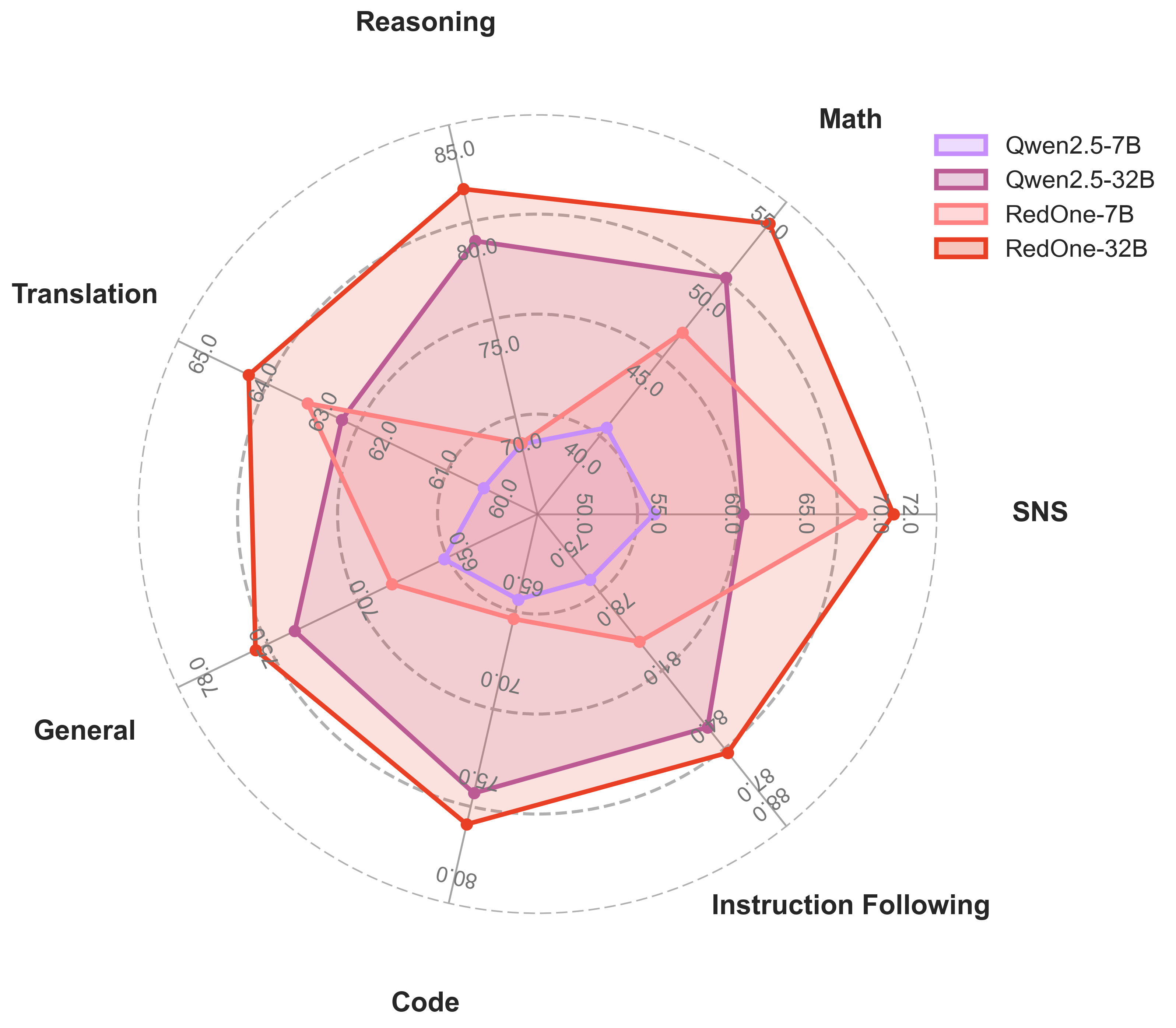}
    \caption{Model capability radar diagram across different task categories.}
    \label{fig:radar}
\end{figure}

As shown in Tables \ref{tab:7b} and \ref{tab:32b}, we conducted a comparison between RedOne with baseline models across various tasks in three categories. Meanwhile, as illustrated in Figure \ref{fig:radar}, we compared RedOne-7B and RedOne-32B with their base model across seven dimensions (six for general and one for SNS). Both results indicate that RedOne in all scales not only maintain robust general capabilities, even surpassing their base model on general tasks, but also exhibit exceptional effectiveness in the SNS domain. Additionally, RedOne achieves performance comparable to significantly larger models across most tasks, with limited improvement opportunities observed only in few areas. The results also demonstrate that scaling up RedOne consistently enhances performance over smaller variants, aligning with established model scaling laws. These findings underscore RedOne's strong potential for further advances through continued increases in model size, as well as its promise for real world application.

\subsection{Task-specific SFT Comparison}
To further explore the impact of base model selection on task-specific fine-tuning and validate our domain LLM's effectiveness, we conducted experiments on two 7B-scale models: the original Qwen-2.5-Instruct (``Qwen'') and our SNS-adapted model (``RedOne''). We evaluated three variants: (1) Qwen-Finetuned, involving task-specific fine-tuning on Qwen; (2) RedOne-Finetuned, involving task-specific fine-tuning on RedOne; and (3) RedOne, representing zero-shot inference without further fine-tuning.

\input{section/task_specific_table}

As shown in Table~\ref{tab:task_specific_sft}, RedOne-Finetuned consistently outperforms Qwen2.5-Finetuned across most datasets, demonstrating that domain-aligned post-training (i.e., RedOne) provides a stronger foundation for downstream SFT. Meanwhile, even RedOne in the zero-shot setting exhibits strong performance, further corroborating the benefits of domain adaptation. Overall, these results indicate that initializing SFT from a domain-adapted base model is more effective than starting from a general-purpose large model. This finding suggests that domain-specific post-training can serve as a powerful approach for improving both zero-shot capabilities and task-specific performance after fine-tuning.

\subsection{Ablation Study}

We also investigate the contributions of each stage in our training pipeline, with results summarized in Table~\ref{tab:ablation}. While CPT alone shows limited immediate gains, it establishes a crucial foundation for subsequent specialization. Based on the CPT model, adding SFT and PO brings average improvements of 0.55 and 1.90 on SNS-Bench compared to variants without CPT, indicating that CPT provides a stronger knowledge foundation for SFT to improve instruction-following and also offers a broader exploration space for PO. Additionally, both CPT and SFT lead to a performance drop on general benchmarks, whereas PO could effectively mitigate this decline and further enhance the overall results. Finally, the complete three-stage pipeline yields the strongest results on specialized benchmarks, with 66.88 on SNS and 48.11 on SNS-Trans, while maintaining competitive general-domain performance at 63.83, demonstrating the effectiveness of our training strategy.

\input{section/ablation_table}

\subsection{Online Results}
\begin{table}[t]
\centering
\resizebox{0.45\textwidth}{!}{
\begin{tabular}{lcc}
\toprule
\textbf{Task} & \textbf{Metric} & \textbf{Change (\%)} \\
\midrule
Harmful Content Detection  & Exposure Rate ($\downarrow$) & $-11.23$ \\
Post-View Search & Click Page Rate ($\uparrow$) & $+14.95$ \\
\bottomrule
\end{tabular}
}
\caption{Effectiveness in online scenarios.}
\label{tab:online_perf}
\end{table}
To further validate RedOne's practical effectiveness, we deployed the model across multiple internal SNS scenarios and witnessed remarkable performance gains in real-world applications compared with previous single-task models as shown in Table~\ref{tab:online_perf}. In harmful content detection, RedOne exhibited exceptional safety capabilities by slashing the exposure rate of harmful notes by 11.23\%, effectively filtering out non-compliant content and strengthening platform security. Moreover, for post-view search recommendation, the model delivered a 14.95\% increase in click page rate, indicating improved content discovery and enhanced user engagement following note interactions. These online results demonstrate the strong practical utility of RedOne in real-world scenarios.

\subsection{Out-of-Domain Ability Analysis}
\begin{figure}[t]
    \centering
    \includegraphics[width=0.4\textwidth]{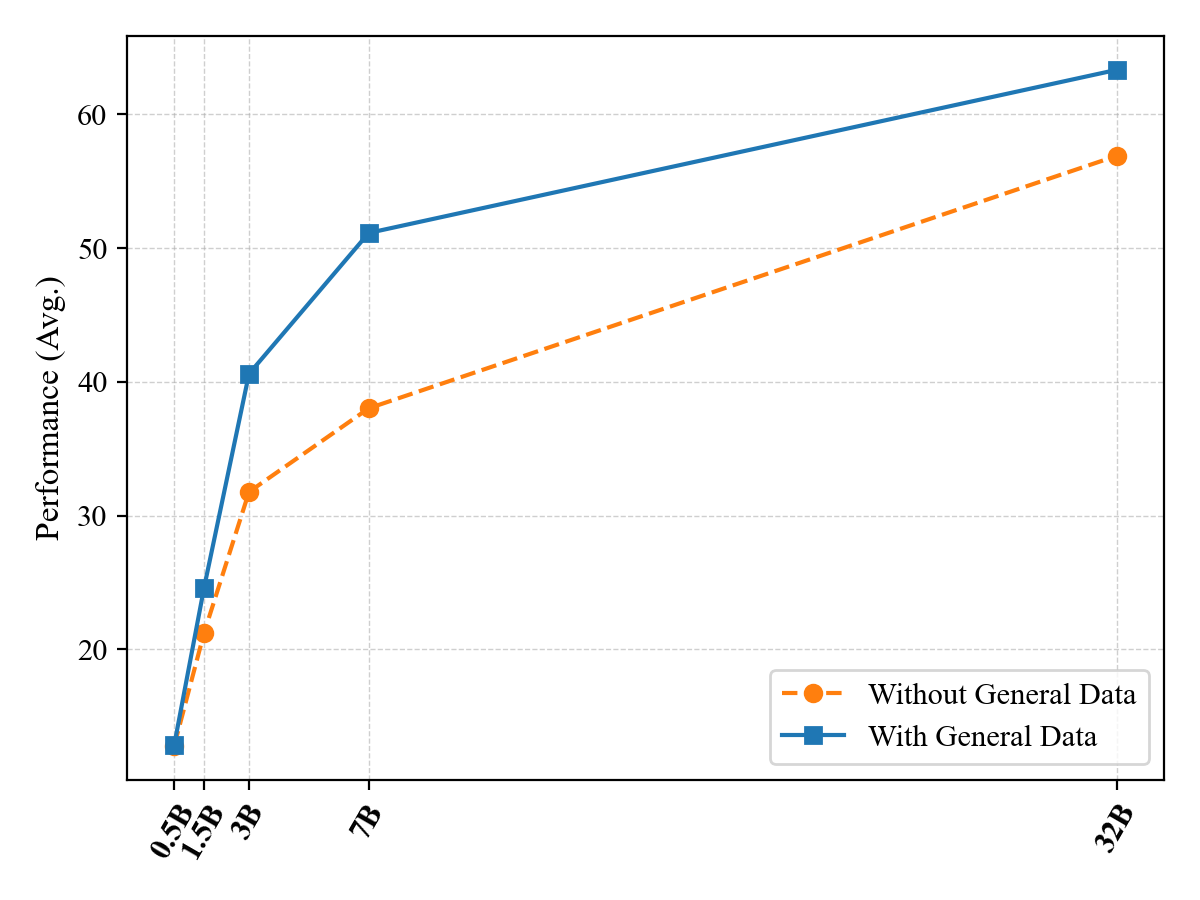}
    \caption{Performance on OOD tasks for models of varying parameter size. }
    \label{fig:ood}
\end{figure}

In this subsection, we examine the impact of preserving general-domain capabilities during domain adaptation by evaluating out-of-domain (OOD) robustness. Specifically, we select one task without corresponding supervised training data (Note Taxonomy) and two tasks with available data (Note Hashtag, Note MRC) from the SNS bench. For the latter two, we remove their related training data during supervised fine-tuning (SFT), effectively making all three tasks OOD. 

We then compare models trained with both general and SNS data against those trained with SNS data only, across various model sizes.
Our results (Figure~\ref{fig:ood}) suggest that including general-domain data helps models generalize better to OOD tasks, and this trend is more visible for larger models. This highlights that maintaining general capabilities can be beneficial for domain adaptation, though further study is needed to fully understand this effect.

\subsection{Case Study}

\begin{table}[t]
    \centering
    \scriptsize
    \begin{tabularx}{0.45\textwidth}{l|X}
        \toprule
        \textbf{Input} & 
        \textbf{Title:} Found it! A Soft-Sole Commuter Loafer You Can Walk In All Day. \\
        & \textbf{Tags:} Height Increasing Thick Sole Shoes, Loafers, Beautiful Loafers. \\
        & \textbf{Popular Comments:} How to buy; link.\\
        \midrule
        \textbf{Qwen} & How to choose height-increasing platform shoes. \\
        \midrule
        \textbf{RedOne} & height-increasing thick-soled loafers. \\
        \bottomrule
    \end{tabularx}
    \caption{Post-view search case: Input SNS context and model-generated queries.}
    \label{tab:post_view_shoes_case}
\end{table}

To further demonstrate RedOne's effectiveness in capturing user search intent within SNS scenarios, we present a case study on the post-view search task. As shown in Table~\ref{tab:post_view_shoes_case}, we analyze a sample post featuring height-increasing loafers that generated significant purchase intent among users. Qwen produces a general shopping query, whereas RedOne directly identifies the core product keywords, better reflecting users' intent to search for and purchase the featured item. This demonstrates RedOne's superior capability for generating actionable queries aligned with real user needs. 

%% file: section/7b_table.tex
\begin{table*}[h]
\centering
\resizebox{1\textwidth}{!}{
\begin{tabular}{l|c|cccccccc|c|cccc|c}
\toprule
\multirow{4}{*}{\textbf{Models}} & 
\multicolumn{1}{c|}{\textbf{General-Bench}} & 
\multicolumn{9}{c|}{\textbf{SNS-Bench}} &
\multicolumn{5}{c}{\textbf{SNS-TransBench}} 
\\
\cmidrule(lr){2-2}
\cmidrule(lr){3-11}
\cmidrule(lr){12-16}
& \multirow{2.5}{*}{\textbf{Avg.}} &
 \multirow{2.5}{*}{\textbf{Taxonomy}} & 
 \multirow{2.5}{*}{\textbf{Hashtag}} & 
 \multirow{2.5}{*}{\textbf{QueryCorr}} & 
 \multirow{2.5}{*}{\textbf{MRC}} &
 \multirow{2.5}{*}{\textbf{NER}} & 
 \multirow{2.5}{*}{\textbf{Gender}} & 
 \multirow{2.5}{*}{\textbf{CHLW}} & 
 \multirow{2.5}{*}{\textbf{QueryGen}} &
 \multirow{2.5}{*}{\textbf{Avg.}} &
 \multicolumn{2}{c}{\textbf{ZH→EN}} & 
 \multicolumn{2}{c|}{\textbf{EN→ZH}} & 
 \multirow{2.5}{*}{\textbf{Avg.}} 
 \\
\cmidrule(lr){12-15}
& & & & & & & & & & &
\textbf{BLEU} & \textbf{chrF++} & \textbf{BLEU} & \textbf{chrF++} &
 \\

\midrule

Llama-3.1-8B \cite{llama3-1} & 51.24 &
37.74 & 66.62 & 33.32 & 31.27 & \underline{47.10} & 74.61 & 26.88 & 38.60 & 44.52 &
23.07 & 48.15 & 29.32 & 29.13 & 32.42 \\
Ministral-8B \cite{ministral} & 49.93 &
42.62 & 70.58 & 36.24 & 30.71 & 37.79 & \underline{82.38} & 28.04 & \underline{46.27} & 46.83 &
25.67 & 50.91 & 32.02 & 31.18 & 34.95 \\
InternLM3-8B \cite{internlm3} & 58.55 &
51.83 & 76.98 & 38.65 & 25.25 & 39.41 & 66.84 & \underline{44.71} & 43.46 & 48.39 & 
24.85 & 50.44 & 35.58 & 34.04 & 36.23 \\

GLM-4-9B-0414 \cite{glm-z1} & \underline{63.27} & 
\underline{56.03} & \underline{77.67} & 38.03 & 45.29 & 47.01 & 51.30 & 27.51 & 45.52 & 48.55 & 
\underline{32.20} & \underline{56.90} & \underline{39.73} & \underline{37.40} & \underline{41.57} \\

\midrule

Qwen2.5-7B \cite{qwen2-5} & 63.01 &
49.50 & 73.80 & \underline{42.37} & \underline{45.32} & 45.41 & \textbf{88.08} & 33.76 & 44.65 & \underline{52.86} &
31.43 & 55.91 & 38.36 & 36.48 & 40.55 \\
RedOne-7B \textbf{(Ours)} & \textbf{63.83} \scriptsize{(+0.82\%)} & 
\textbf{72.18} & \textbf{88.02} & \textbf{65.09} & \textbf{63.98} & \textbf{51.86} & 70.47 & \textbf{74.73} & \textbf{48.69} & \textbf{66.88} \scriptsize{(+14.02\%)} & 
\textbf{38.06} & \textbf{62.66} & \textbf{46.88} & \textbf{44.82} & \textbf{48.11} \scriptsize{(+7.56\%)} \\

\bottomrule
\end{tabular}%
}

\caption{
Results of 7B-scale models.
\textbf{Bold} entries indicate the best model, while \underline{underlined} entries denote the second one. Percentage improvements relative to the baseline Qwen2.5 foundation model are also shown.
}
\label{tab:7b}
\end{table*}

%% file: section/32b_table.tex
\begin{table*}[h]
\centering
\resizebox{1\textwidth}{!}{
\begin{tabular}{l|c|cccccccc|c|cccc|c}
\toprule
\multirow{3}{*}{\textbf{Models}} & 
\multicolumn{1}{c|}{\textbf{General-Bench}} & 
\multicolumn{9}{c|}{\textbf{SNS-Bench}} &
\multicolumn{5}{c}{\textbf{SNS-TransBench}} 
\\
\cmidrule(lr){2-2}
\cmidrule(lr){3-11}
\cmidrule(lr){12-16}
& \multirow{2.5}{*}{\textbf{Avg.}} &
 \multirow{2.5}{*}{\textbf{Taxonomy}} & 
 \multirow{2.5}{*}{\textbf{Hashtag}} & 
 \multirow{2.5}{*}{\textbf{QueryCorr}} & 
 \multirow{2.5}{*}{\textbf{MRC}} &
 \multirow{2.5}{*}{\textbf{NER}} & 
 \multirow{2.5}{*}{\textbf{Gender}} & 
 \multirow{2.5}{*}{\textbf{CHLW}} & 
 \multirow{2.5}{*}{\textbf{QueryGen}} &
 \multirow{2.5}{*}{\textbf{Avg.}} &
 \multicolumn{2}{c}{\textbf{ZH→EN}} & 
 \multicolumn{2}{c|}{\textbf{EN→ZH}} & 
 \multirow{2.5}{*}{\textbf{Avg.}} 
 \\
\cmidrule(lr){12-15}
& & & & & & & & & & &
\textbf{BLEU} & \textbf{chrF++} & \textbf{BLEU} & \textbf{chrF++} &
 \\
\midrule
\multicolumn{16}{c}{\textit{Open-Source Large Language Models}} \\
\midrule

Phi-4-14B \cite{phi4-14} & 63.00 &
57.62 & 79.56 & 46.32 & 53.39 & 44.99 & 89.12 & 29.23 & 44.76 & 55.62 & 
31.28 & 57.23 & 37.58 & 36.68 & 40.69 \\
Mistral-Small-24B \cite{mistralsmall2025}  & 65.63 &
64.88 & 83.89 & 48.77 & 46.51 & 52.09 & \underline{91.19} & 32.10 & 46.01 & 58.18 & 
31.29 & 56.72 & 39.28 & 37.32 & 41.15 \\
Llama-3.3-70B \cite{llama3-1} & 67.64 &
62.94 & 83.28 & 50.76 & 27.38 & 56.09 & \underline{91.19} & 33.58 & 46.41 & 56.45 & 
34.00 & 59.18 & 41.25 & 39.56 & 43.50 \\
GLM-4-32B-0414 \cite{glm-z1} & 74.39 & 
63.36 & 85.50 & 47.33 & 53.72 & 50.41 & 80.31 & 33.19 & 46.90 & 57.59 & 
36.32 & 61.31 & 42.53 & 40.77 & 45.23 \\
Deepseek-V3-0324 \cite{deepseekv3} & \underline{75.22} &
67.27 & 86.59 & 47.71 & 60.97 & 56.00 & 90.16 & \underline{40.45} & 46.03 & 61.90 & 
35.65 & 61.58 & 46.86 & 44.58 & 47.17 \\

% ==========================================
\midrule
\multicolumn{16}{c}{\textit{Closed-Source Large Language Models}} \\
\midrule

Doubao-1.5-Pro-32k \cite{doubao_1_5_pro} & \textbf{76.13} &
\textit{30.00} & 83.21 & \underline{58.25} & \underline{61.32} & \textbf{56.60} & 90.67 & 30.61 & 46.55 & 57.15 & 
33.71 & 61.85 & 45.54 & 44.35 & 46.36 \\
GLM-4-Plus \cite{glm-z1} & \textit{70.25} &
65.46 & 84.31 & 52.13 & 55.81 & 53.16 & 86.53 & 30.09 & 44.68 & 59.02 & 
\textbf{41.57} & \textbf{65.95} & \underline{48.79} & \underline{47.06} & \textbf{50.84} \\
GPT-4o-1120 \cite{gpt4o-1120} & \textit{70.72} &
65.79 & 84.98 & 51.79 & 58.89 & 54.99 & 88.08 & 38.96 & \underline{47.33} & 61.35 & 
40.32 & 63.91 & \textbf{49.15} & \textbf{47.28} & \underline{50.17} \\
Claude-3.7-Sonnet \cite{claude3-7} & 75.10 &
\underline{72.03} & \underline{88.83} & 54.10 & 54.86 & \underline{56.13} & \textbf{92.23} & 31.11 & 45.49 & 61.85 & 
35.63 & 61.66 & 45.79 & 44.23 & 46.83 \\
Gemini-2.0-Flash \cite{gemini2} & 74.42 & 
68.76 & 87.36 & 48.41 & 52.21 & 53.58 & 89.64 & 37.39 & 46.27 & 60.45 & 
32.72 & 58.84 & 41.80 & 40.16 & 43.38 \\
Qwen-Max \cite{qwen2-5} & 71.86 &
65.68 & 84.47 & 54.36 & \textbf{61.34} & 55.78 & \underline{91.19} & 37.97 & 46.64 & \underline{62.18} & 
35.55 & 60.92 & 46.08 & 44.14 & 46.67 \\

\midrule

Qwen2.5-32B \cite{qwen2-5} & 71.68 &
59.90 & 80.51 & 46.00 & 55.04 & 54.51 & 90.67 & 38.84 & 45.66 & 58.89 & 
32.56 & 58.14 & 42.34 & 40.71 & 43.44 \\

RedOne-32B \textbf{(Ours)} & 73.72 \scriptsize{(+2.04\%)} & 
\textbf{81.45} & \textbf{90.19} & \textbf{67.07} & 59.24 & 51.66 & 81.87 & \textbf{70.40} & \textbf{50.37} & \textbf{69.03} \scriptsize{(+10.14\%)} & 
\underline{40.55} & \underline{64.54} & 48.20 & 46.05 & 49.84 \scriptsize{(+6.40\%)} \\

\bottomrule
\end{tabular}%
}
\caption{Results of 32B-scale models. \textbf{Bold} entries indicate the best model, while \underline{underlined} entries denote the second one. Percentage improvements relative to the baseline Qwen2.5 foundation model are also shown.}
\label{tab:32b}
\end{table*}

%% file: section/task_specific_table.tex
\begin{table}[t]
\centering
\resizebox{0.45\textwidth}{!}{
\begin{tabular}{lccc}
\toprule
\textbf{Models} & \textbf{HashTag} & \textbf{QueryCorr} & \textbf{MRC} \\
\midrule
Qwen2.5-Finetuned      & 88.93 & 57.76 & 62.26 \\
RedOne & 88.02 \scriptsize{(-0.91\%)} & 65.09 \scriptsize{(+12.63\%)} & 63.98 \scriptsize{(+2.76\%)} \\
% RedOne        & 88.02 & 65.09 & 63.98 \\
RedOne-Finetuned    & \textbf{90.51}\scriptsize{(+1.78\%)} & \textbf{65.77}\scriptsize{(+13.87\%)} & \textbf{64.47}\scriptsize{(+3.55\%)} \\
\midrule
\textbf{Models} & \textbf{CHLW} & \textbf{QueryGen} & \textbf{SNS-Trans} \\
\midrule
Qwen2.5-Finetuned      & 78.41 & 48.25 & 48.01 \\
% RedOne          & 74.73 & 48.69 & 48.11 \\
RedOne & 74.73 \scriptsize{(-4.72\%)} & 48.69 \scriptsize{(+0.91\%)} & 48.11 \scriptsize{(+0.21\%)} \\
RedOne-Finetuned    & \textbf{79.11}\scriptsize{(+0.89\%)} & \textbf{49.21}\scriptsize{(+1.99\%)} & \textbf{48.32}\scriptsize{(+0.65\%)} \\
\bottomrule
\end{tabular}
}
\caption{Performance comparison of task-specific Finetuned on Qwen-2.5-Instruct and RedOne (all models are 7B scale).}
\label{tab:task_specific_sft}
\end{table}

%% file: section/ablation_table.tex
\begin{table}[t]
\centering

\resizebox{0.4\textwidth}{!}{
\begin{tabular}{ccc|ccc}
\toprule
\textbf{CPT} & \textbf{SFT} & \textbf{PO} & 
\multicolumn{1}{c}{\textbf{General}} & 
\multicolumn{1}{c}{\textbf{SNS}} &
\multicolumn{1}{c}{\textbf{SNS-Trans}} 
 \\
% ==========================================
\midrule
 & & & 63.01 & 52.86 & 40.55 \\
 & {$\checkmark$} & & 62.65 & 
 64.57 & 
 47.47 \\
 & {$\checkmark$} & {$\checkmark$} & \textbf{64.36} & 
64.98 & 
47.64 \\
{$\checkmark$} & & & 62.28 & 
53.28 & 
41.39 \\
{$\checkmark$} & {$\checkmark$} & & 61.95 & 
65.12 & 
47.70 \\
{$\checkmark$} & {$\checkmark$} & {$\checkmark$} & 63.83 & 
\textbf{66.88} & 
\textbf{48.11} \\
\bottomrule
\end{tabular}%
}
\caption{
Ablation study results of RedOne-7B.
}

\label{tab:ablation}
\end{table}

%% file: section/conclusion.tex
\section{Conclusion}
In this paper, we introduce RedOne, a domain-specific LLM trained through a three-stage strategy that enhances SNS-specific capabilities while preserving general performance. We believe our approach can inspire future research in developing specialized LLMs and advancing practical applications in social media.

\section*{Limitations}
Although our proposed method demonstrates strong effectiveness, it requires extensive data processing, resulting in considerable resource costs. Additionally, current models are still in a large scale, which increases online inference latency and serving expenses, limiting deployment in resource-constrained settings. Future work will explore lighter architectures, including model compression through quantization and distillation, as well as routing-efficient designs such as mixture-of-experts (MoE), to reduce latency and cost without sacrificing accuracy.

\section*{Ethical Considerations}
When integrating large language models as essential components within application services, it is crucial to rigorously consider potential model hallucinations and security risks. To ensure reliable service delivery, leveraging RedOne for model services requires careful implementation to mitigate adverse user impacts. Furthermore, we emphasize the critical importance of adhering to stringent user privacy protection standards throughout data collection and processing workflows, ensuring comprehensive personal information security.

\section*{Reproducibility}
Due to data privacy and code security concerns, we are currently unable to fully release our datasets and code. However, our entire development pipeline is built upon widely adopted open-source projects and works~\cite{zheng2024llamafactory,rafailov2024direct,dong2023abilities}, which ensures that users only need to organize their data according to the specified format in order to run it smoothly. In addition, we have reported detailed parameter settings used during training in Section~\ref{sec:imp_details}, which further provides valuable references for the community to reproduce our pipeline.

%% file: section/appendix.tex
\section{Appendices}
\subsection{SNS Tasks}
\label{appendix:sns_tasks}

In this section, we provide an overview of the key tasks defined for SFT in SNS scenarios. These tasks are designed to reflect real-world user behavior, content patterns throughout social platforms. The SFT task suite covers six core capability areas, each capturing an important aspect of SNS applications:
\paragraph{Content Understanding:} This category focuses on the model's ability to comprehend and categorize user-generated content as well as user queries. Example tasks include classifying notes into categories (\emph{note taxonomy}), determining the topic or domain of user queries (\emph{query classification}), and identifying fine-grained query intent (\emph{query intent recognition}).
\paragraph{Information Extraction:} Tasks in this category address the identification and extraction of structured information from informal SNS posts. This includes predicting appropriate hashtags for a post, answering questions about note content, and detecting highlight or anchor words that represent user focus.
\paragraph{Semantic Matching:} Here, the model is required to judge the semantic relationship and relevance between items such as user queries and social notes. Typical tasks include evaluating whether a note is relevant to a given query (\emph{query-note relevance}) and retrieving the most pertinent or high-quality notes for search scenarios (\emph{query-note retrieval}).
\paragraph{User Behavior Modeling:} This capability involves modeling and simulating user actions, such as generating follow-up queries based on previous browsing or posting activities (\emph{post-view search}). It reflects how users might interact with content in a dynamic SNS environment.
\paragraph{Dialogue and Persona Simulation:} To enhance natural interaction and personalization, dialogue tasks ask the model to engage in emotional companion conversations or role-play as different personas in group chats, capturing both the style and richness of real SNS dialogues.
\paragraph{Translation:} Given the prevalence of multilingual content, the model is also trained to translate notes between languages, with attention to preserving the original tone, sentiment, and informal expressions common across SNS platforms.

Each task adopts the most suitable instruction-tuning format: multiple choice supports classification and selection, extraction is used for entity and span prediction, and generation handles open-ended responses such as dialogue or translation. This format-driven design ensures consistent prompting and facilitates efficient multi-task training.

\begin{figure}[!h]
\centering
\includegraphics[width=0.48\textwidth]{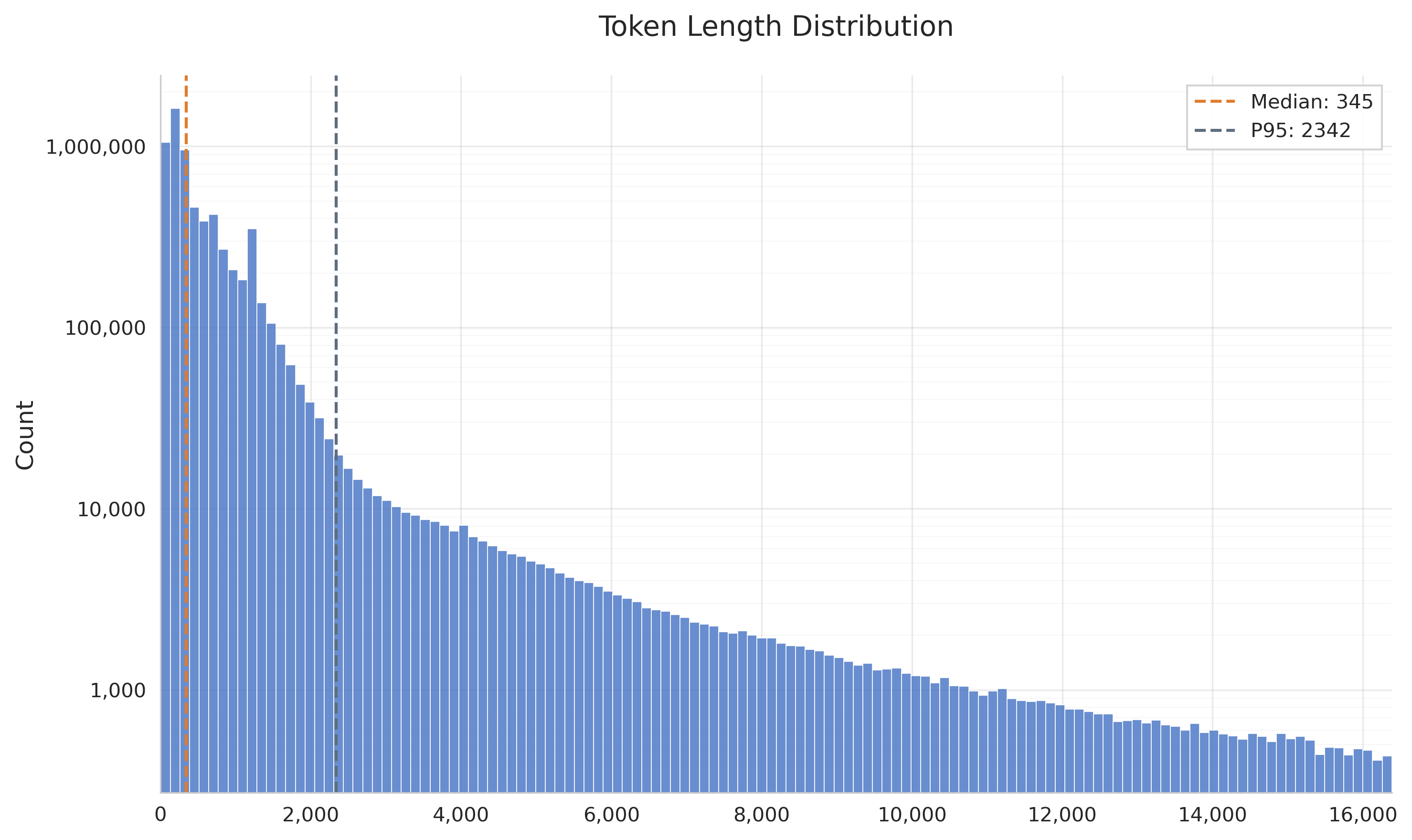}
\caption{Token length distribution in the dataset. The histogram uses a logarithmic y-axis with dashed lines indicating the median (345 tokens) and the 95-th percentile (2,342 tokens).}
\label{fig:token_length_dist}
\end{figure}

\begin{figure}[!h]
    \centering
    \begin{subfigure}[t]{0.48\textwidth}
        \centering
        \includegraphics[width=\textwidth]{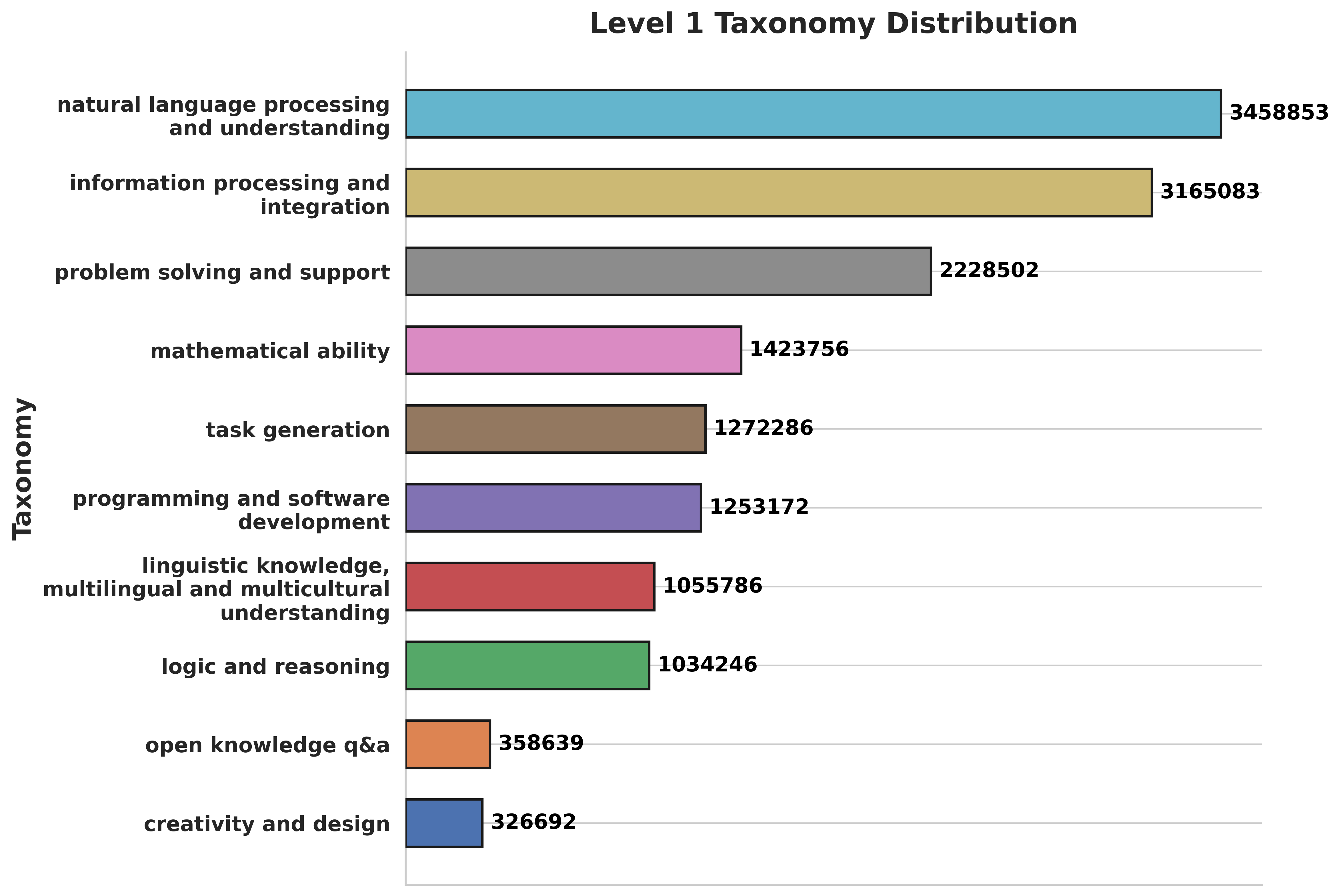}
        \caption{Primary task categories.}
        \label{fig:primary_label_dist}
    \end{subfigure}
    \hfill
    \begin{subfigure}[t]{0.48\textwidth}
        \centering
        \includegraphics[width=\textwidth]{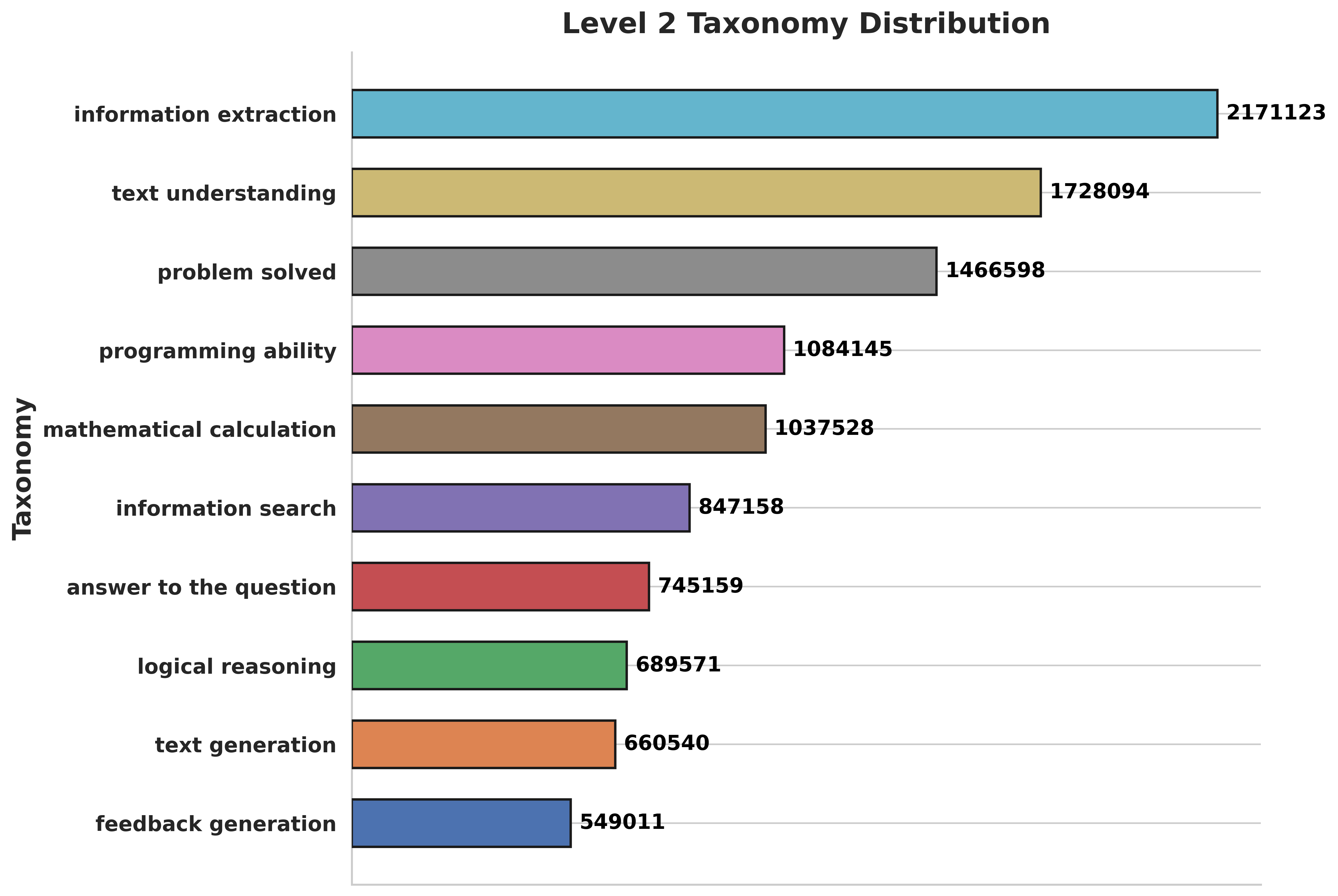}
        \caption{Secondary task categories.}
        \label{fig:secondary_label_dist}
    \end{subfigure}
    \caption{Top 10 distributions of primary and secondary task categories in the SFT dataset.}
    \label{fig:taxonomy_distributions}
\end{figure}

\subsection{SFT Data Statistical Analysis}
\label{appendix:sft_data_statistics}

\paragraph{Token Length Statistics}
Figure~\ref{fig:token_length_dist} presents the token length distribution across all samples in our dataset, displayed on a logarithmic scale to accommodate the wide range of sequence lengths up to 16,384 tokens. The distribution exhibits the characteristic heavy-tailed pattern typical of natural language corpora.

\paragraph{Task Category Distribution}
We adopted the labeling taxonomy from Infinity Instruct~\cite{li2025infinity}, organizing it into primary and secondary categories. Specifically, we used a subset of Infinity Instruct data, consisting of instructions paired with their corresponding labels, as training data to fine-tune a labeling model based on Qwen2.5-7B-Instruct. This trained labeling model was then applied to annotate all instructions in our complete SFT dataset. The comprehensive distribution of primary and secondary label categories is illustrated in Figures~\ref{fig:primary_label_dist} and~\ref{fig:secondary_label_dist}, respectively.

The distribution analysis reveals several key characteristics of our SFT dataset. At the primary level, natural language processing and understanding dominates with over 3.4 million instances, followed closely by information processing and integration (3.1 million) and problem solving and support (2.2 million). This indicates a strong emphasis on core language comprehension and analytical capabilities. Mathematical ability and programming-related tasks also constitute significant portions, with over 1.2 million instances each, reflecting the dataset's comprehensive coverage of technical skills. At the secondary level, information extraction leads with 2.1 million instances, while text understanding and problem-solving tasks follow with 1.7 million and 1.4 million instances respectively. The relatively balanced distribution across different cognitive abilities suggests that our dataset provides diverse training scenarios for developing well-rounded AI capabilities.

%% file: acl_latex.bbl
\begin{thebibliography}{88}
\providecommand{\natexlab}[1]{#1}

\bibitem[{Abdin et~al.(2024)Abdin, Aneja, Behl, Bubeck, Eldan, Gunasekar, Harrison, Hewett, Javaheripi, Kauffmann, Lee, Lee, Li, Liu, Mendes, Nguyen, Price, de~Rosa, Saarikivi, Salim, Shah, Wang, Ward, Wu, Yu, Zhang, and Zhang}]{phi4-14}
Marah Abdin, Jyoti Aneja, Harkirat Behl, Sébastien Bubeck, Ronen Eldan, Suriya Gunasekar, Michael Harrison, Russell~J. Hewett, Mojan Javaheripi, Piero Kauffmann, James~R. Lee, Yin~Tat Lee, Yuanzhi Li, Weishung Liu, Caio C.~T. Mendes, Anh Nguyen, Eric Price, Gustavo de~Rosa, Olli Saarikivi, and 8 others. 2024.
\newblock \href {https://arxiv.org/abs/2412.08905} {Phi-4 technical report}.
\newblock \emph{Preprint}, arXiv:2412.08905.

\bibitem[{Anthropic()}]{claude3-7}
Anthropic.
\newblock Claude 3.7 sonnet and claude code.
\newblock \url{https://www.anthropic.com/news/claude-3-7-sonnet}.

\bibitem[{Austin et~al.(2021)Austin, Odena, Nye, Bosma, Michalewski, Dohan, Jiang, Cai, Terry, Le, and Sutton}]{mbpp}
Jacob Austin, Augustus Odena, Maxwell~I. Nye, Maarten Bosma, Henryk Michalewski, David Dohan, Ellen Jiang, Carrie~J. Cai, Michael Terry, Quoc~V. Le, and Charles Sutton. 2021.
\newblock Program synthesis with large language models.
\newblock \emph{CoRR}, abs/2108.07732.

\bibitem[{Azerbayev et~al.(2023)Azerbayev, Schoelkopf, Paster, Santos, McAleer, Jiang, Deng, Biderman, and Welleck}]{azerbayev2023llemma}
Zhangir Azerbayev, Hailey Schoelkopf, Keiran Paster, Marco~Dos Santos, Stephen McAleer, Albert~Q Jiang, Jia Deng, Stella Biderman, and Sean Welleck. 2023.
\newblock Llemma: An open language model for mathematics.
\newblock \emph{arXiv preprint arXiv:2310.10631}.

\bibitem[{Bi et~al.(2023)Bi, Zhang, Xue, Ou, Ji, Zheng, and Chen}]{bi2023oceangpt}
Zhen Bi, Ningyu Zhang, Yida Xue, Yixin Ou, Daxiong Ji, Guozhou Zheng, and Huajun Chen. 2023.
\newblock Oceangpt: A large language model for ocean science tasks.
\newblock \emph{arXiv preprint arXiv:2310.02031}.

\bibitem[{Cai et~al.(2024)Cai, Cao, Chen, Chen, Chen, Chen, Chen, Chen, Chen, Chu, Dong, Duan, Fan, Fei, Gao, Ge, Gu, Gu, Gui, Guo, Guo, He, Hu, Huang, Jiang, Jiao, Jin, Lei, Li, Li, Li, Li, Li, Li, Liu, Liu, Hong, Liu, Liu, Liu, Lv, Lv, Lv, Ma, Ma, Ma, Ning, Ouyang, Qiu, Qu, Shang, Shao, Song, Song, Sui, Sun, Sun, Tang, Wang, Wang, Wang, Wang, Wang, Wang, Wang, Wei, Weng, Wu, Xiong, Xu, Xu, Yan, Yan, Yang, Ye, Ying, Yu, Yu, Zang, Zhang, Zhang, Zhang, Zhang, Zhang, Zhang, Zhang, Zhang, Zhang, Zhang, Zhang, Zhao, Zhao, Zhao, Zhou, Zhou, Zhuo, Zou, Qiu, Qiao, and Lin}]{internlm3}
Zheng Cai, Maosong Cao, Haojiong Chen, Kai Chen, Keyu Chen, Xin Chen, Xun Chen, Zehui Chen, Zhi Chen, Pei Chu, Xiaoyi Dong, Haodong Duan, Qi~Fan, Zhaoye Fei, Yang Gao, Jiaye Ge, Chenya Gu, Yuzhe Gu, Tao Gui, and 81 others. 2024.
\newblock \href {https://arxiv.org/abs/2403.17297} {Internlm2 technical report}.
\newblock \emph{Preprint}, arXiv:2403.17297.

\bibitem[{Cao et~al.(2024{\natexlab{a}})Cao, Lam, Duan, Liu, Zhang, and Chen}]{cao2024compassjudger}
Maosong Cao, Alexander Lam, Haodong Duan, Hongwei Liu, Songyang Zhang, and Kai Chen. 2024{\natexlab{a}}.
\newblock Compassjudger-1: All-in-one judge model helps model evaluation and evolution.
\newblock \emph{arXiv preprint arXiv:2410.16256}.

\bibitem[{Cao et~al.(2024{\natexlab{b}})Cao, Lam, Duan, Liu, Zhang, and Chen}]{compassbench}
Maosong Cao, Alexander Lam, Haodong Duan, Hongwei Liu, Songyang Zhang, and Kai Chen. 2024{\natexlab{b}}.
\newblock \href {https://arxiv.org/abs/2410.16256} {Compassjudger-1: All-in-one judge model helps model evaluation and evolution}.
\newblock abs/2410.16256.

\bibitem[{Carr and Hayes(2015)}]{carr2015social}
Caleb~T Carr and Rebecca~A Hayes. 2015.
\newblock Social media: Defining, developing, and divining.
\newblock \emph{Atlantic journal of communication}, 23(1):46--65.

\bibitem[{Chen et~al.(2021)Chen, Tworek, Jun, Yuan, de~Oliveira~Pinto, Kaplan, Edwards, Burda, Joseph, Brockman, Ray, Puri, Krueger, Petrov, Khlaaf, Sastry, Mishkin, Chan, Gray, Ryder, Pavlov, Power, Kaiser, Bavarian, Winter, Tillet, Such, Cummings, Plappert, Chantzis, Barnes, Herbert{-}Voss, Guss, Nichol, Paino, Tezak, Tang, Babuschkin, Balaji, Jain, Saunders, Hesse, Carr, Leike, Achiam, Misra, Morikawa, Radford, Knight, Brundage, Murati, Mayer, Welinder, McGrew, Amodei, McCandlish, Sutskever, and Zaremba}]{humaneval}
Mark Chen, Jerry Tworek, Heewoo Jun, Qiming Yuan, Henrique~Pond{\'{e}} de~Oliveira~Pinto, Jared Kaplan, Harrison Edwards, Yuri Burda, Nicholas Joseph, Greg Brockman, Alex Ray, Raul Puri, Gretchen Krueger, Michael Petrov, Heidy Khlaaf, Girish Sastry, Pamela Mishkin, Brooke Chan, Scott Gray, and 39 others. 2021.
\newblock Evaluating large language models trained on code.
\newblock \emph{CoRR}, abs/2107.03374.

\bibitem[{Chen et~al.(2023)Chen, Wang, Xing, Xu, Fang, Wang, Li, Wu, Liu, Xu et~al.}]{chen2023bianque}
Yirong Chen, Zhenyu Wang, Xiaofen Xing, Zhipei Xu, Kai Fang, Junhong Wang, Sihang Li, Jieling Wu, Qi~Liu, Xiangmin Xu, and 1 others. 2023.
\newblock Bianque: Balancing the questioning and suggestion ability of health llms with multi-turn health conversations polished by chatgpt.
\newblock \emph{arXiv preprint arXiv:2310.15896}.

\bibitem[{Chu et~al.(2025)Chu, Zhai, Yang, Tong, Xie, Schuurmans, Le, Levine, and Ma}]{chu2025sft}
Tianzhe Chu, Yuexiang Zhai, Jihan Yang, Shengbang Tong, Saining Xie, Dale Schuurmans, Quoc~V Le, Sergey Levine, and Yi~Ma. 2025.
\newblock Sft memorizes, rl generalizes: A comparative study of foundation model post-training.
\newblock \emph{arXiv preprint arXiv:2501.17161}.

\bibitem[{Cobbe et~al.(2021)Cobbe, Kosaraju, Bavarian, Chen, Jun, Kaiser, Plappert, Tworek, Hilton, Nakano, Hesse, and Schulman}]{gsm8k}
Karl Cobbe, Vineet Kosaraju, Mohammad Bavarian, Mark Chen, Heewoo Jun, Lukasz Kaiser, Matthias Plappert, Jerry Tworek, Jacob Hilton, Reiichiro Nakano, Christopher Hesse, and John Schulman. 2021.
\newblock Training verifiers to solve math word problems.
\newblock \emph{CoRR}, abs/2110.14168.

\bibitem[{Colombo et~al.(2024)Colombo, Pires, Boudiaf, Culver, Melo, Corro, Martins, Esposito, Raposo, Morgado et~al.}]{colombo2024saullm}
Pierre Colombo, Telmo~Pessoa Pires, Malik Boudiaf, Dominic Culver, Rui Melo, Caio Corro, Andre~FT Martins, Fabrizio Esposito, Vera~L{\'u}cia Raposo, Sofia Morgado, and 1 others. 2024.
\newblock Saullm-7b: A pioneering large language model for law.
\newblock \emph{arXiv preprint arXiv:2403.03883}.

\bibitem[{DeepMind(2024)}]{gemini2}
Google DeepMind. 2024.
\newblock Introducing gemini 2.0: our new ai model for the agentic era.
\newblock \url{https://blog.google/technology/google-deepmind/google-gemini-ai-update-december-2024/}.

\bibitem[{DeepSeek-AI et~al.(2025)DeepSeek-AI, Liu, Feng, Xue, Wang, Wu, Lu, Zhao, Deng, Zhang, Ruan, Dai, Guo, Yang, Chen, Ji, Li, Lin, Dai, Luo, Hao, Chen, Li, Zhang, Bao, Xu, Wang, Zhang, Ding, Xin, Gao, Li, Qu, Cai, Liang, Guo, Ni, Li, Wang, Chen, Chen, Yuan, Qiu, Li, Song, Dong, Hu, Gao, Guan, Huang, Yu, Wang, Zhang, Xu, Xia, Zhao, Wang, Zhang, Li, Wang, Zhang, Zhang, Tang, Li, Tian, Huang, Wang, Zhang, Wang, Zhu, Chen, Du, Chen, Jin, Ge, Zhang, Pan, Wang, Xu, Zhang, Chen, Li, Lu, Zhou, Chen, Wu, Ye, Ye, Ma, Wang, Zhou, Yu, Zhou, Pan, Wang, Yun, Pei, Sun, Xiao, Zeng, Zhao, An, Liu, Liang, Gao, Yu, Zhang, Li, Jin, Wang, Bi, Liu, Wang, Shen, Chen, Zhang, Chen, Nie, Sun, Wang, Cheng, Liu, Xie, Liu, Yu, Song, Shan, Zhou, Yang, Li, Su, Lin, Li, Wang, Wei, Zhu, Zhang, Xu, Xu, Huang, Li, Zhao, Sun, Li, Wang, Yu, Zheng, Zhang, Shi, Xiong, He, Tang, Piao, Wang, Tan, Ma, Liu, Guo, Wu, Ou, Zhu, Wang, Gong, Zou, He, Zha, Xiong, Ma, Yan, Luo, You, Liu, Zhou, Wu, Ren, Ren, Sha, Fu, Xu, Huang, Zhang, Xie, Zhang, Hao,
  Gou, Ma, Yan, Shao, Xu, Wu, Zhang, Li, Gu, Zhu, Liu, Li, Xie, Song, Gao, and Pan}]{deepseekv3}
DeepSeek-AI, Aixin Liu, Bei Feng, Bing Xue, Bingxuan Wang, Bochao Wu, Chengda Lu, Chenggang Zhao, Chengqi Deng, Chenyu Zhang, Chong Ruan, Damai Dai, Daya Guo, Dejian Yang, Deli Chen, Dongjie Ji, Erhang Li, Fangyun Lin, Fucong Dai, and 181 others. 2025.
\newblock \href {https://arxiv.org/abs/2412.19437} {Deepseek-v3 technical report}.
\newblock \emph{Preprint}, arXiv:2412.19437.

\bibitem[{Dong et~al.(2023)Dong, Yuan, Lu, Li, Xue, Liu, Wang, Yuan, Zhou, and Zhou}]{dong2023abilities}
Guanting Dong, Hongyi Yuan, Keming Lu, Chengpeng Li, Mingfeng Xue, Dayiheng Liu, Wei Wang, Zheng Yuan, Chang Zhou, and Jingren Zhou. 2023.
\newblock How abilities in large language models are affected by supervised fine-tuning data composition.
\newblock \emph{CoRR}, abs/2310.05492.

\bibitem[{Dong et~al.(2024)Dong, Yuan, Lu, Li, Xue, Liu, Wang, Yuan, Zhou, and Zhou}]{dong-etal-2024-abilities}
Guanting Dong, Hongyi Yuan, Keming Lu, Chengpeng Li, Mingfeng Xue, Dayiheng Liu, Wei Wang, Zheng Yuan, Chang Zhou, and Jingren Zhou. 2024.
\newblock \href {https://doi.org/10.18653/v1/2024.acl-long.12} {How abilities in large language models are affected by supervised fine-tuning data composition}.
\newblock In \emph{Proceedings of the 62nd Annual Meeting of the Association for Computational Linguistics (Volume 1: Long Papers)}, pages 177--198, Bangkok, Thailand. Association for Computational Linguistics.

\bibitem[{Doubao-Team(2025)}]{doubao_1_5_pro}
Doubao-Team. 2025.
\newblock Doubao-1.5-pro: Model release.
\newblock \url{https://team.doubao.com/en/special/doubao_1_5_pro}.

\bibitem[{Eisenstein(2013)}]{eisenstein2013bad}
Jacob Eisenstein. 2013.
\newblock What to do about bad language on the internet.
\newblock In \emph{Proceedings of the 2013 conference of the North American Chapter of the association for computational linguistics: Human language technologies}, pages 359--369.

\bibitem[{Elahimanesh et~al.(2025)Elahimanesh, Mohammadkhani, and Kasaei}]{elahimanesh2025emotion}
Sina Elahimanesh, Mohammadali Mohammadkhani, and Shohreh Kasaei. 2025.
\newblock Emotion alignment: Discovering the gap between social media and real-world sentiments in persian tweets and images.
\newblock \emph{arXiv preprint arXiv:2504.10662}.

\bibitem[{GLM et~al.(2024)GLM, Zeng, Xu, Wang, Zhang, Yin, Rojas, Feng, Zhao, Lai, Yu, Wang, Sun, Zhang, Cheng, Gui, Tang, Zhang, Li, Zhao, Wu, Zhong, Liu, Huang, Zhang, Zheng, Lu, Duan, Zhang, Cao, Yang, Tam, Zhao, Liu, Xia, Zhang, Gu, Lv, Liu, Liu, Yang, Song, Zhang, An, Xu, Niu, Yang, Li, Bai, Dong, Qi, Wang, Yang, Du, Hou, and Wang}]{glm-z1}
Team GLM, Aohan Zeng, Bin Xu, Bowen Wang, Chenhui Zhang, Da~Yin, Diego Rojas, Guanyu Feng, Hanlin Zhao, Hanyu Lai, Hao Yu, Hongning Wang, Jiadai Sun, Jiajie Zhang, Jiale Cheng, Jiayi Gui, Jie Tang, Jing Zhang, Juanzi Li, and 37 others. 2024.
\newblock \href {https://arxiv.org/abs/2406.12793} {Chatglm: A family of large language models from glm-130b to glm-4 all tools}.

\bibitem[{Goyal et~al.(2022)Goyal, Gao, Chaudhary, Chen, Wenzek, Ju, Krishnan, Ranzato, Guzm{\'{a}}n, and Fan}]{flores}
Naman Goyal, Cynthia Gao, Vishrav Chaudhary, Peng{-}Jen Chen, Guillaume Wenzek, Da~Ju, Sanjana Krishnan, Marc'Aurelio Ranzato, Francisco Guzm{\'{a}}n, and Angela Fan. 2022.
\newblock The {Flores-101} evaluation benchmark for low-resource and multilingual machine translation.
\newblock \emph{Trans. Assoc. Comput. Linguistics}, 10:522--538.

\bibitem[{Grattafiori et~al.(2024)Grattafiori, Dubey, Jauhri, Pandey, Kadian, Al-Dahle, Letman, Mathur, Schelten, Vaughan, Yang, Fan, Goyal, Hartshorn, Yang, Mitra, Sravankumar, Korenev, Hinsvark, Rao, Zhang, Rodriguez, Gregerson, Spataru, Roziere, Biron, Tang, Chern, Caucheteux, Nayak, Bi, Marra, McConnell, Keller, Touret, Wu, Wong, Ferrer, Nikolaidis, Allonsius, Song, Pintz, Livshits, Wyatt, Esiobu, Choudhary, Mahajan, Garcia-Olano, Perino, Hupkes, Lakomkin, AlBadawy, Lobanova, Dinan, Smith, Radenovic, Guzmán, Zhang, Synnaeve, Lee, Anderson, Thattai, Nail, Mialon, Pang, Cucurell, Nguyen, Korevaar, Xu, Touvron, Zarov, Ibarra, Kloumann, Misra, Evtimov, Zhang, Copet, Lee, Geffert, Vranes, Park, Mahadeokar, Shah, van~der Linde, Billock, Hong, Lee, Fu, Chi, Huang, Liu, Wang, Yu, Bitton, Spisak, Park, Rocca, Johnstun, Saxe, Jia, Alwala, Prasad, Upasani, Plawiak, Li, Heafield, Stone, El-Arini, Iyer, Malik, Chiu, Bhalla, Lakhotia, Rantala-Yeary, van~der Maaten, Chen, Tan, Jenkins, Martin, Madaan, Malo, Blecher,
  Landzaat, de~Oliveira, Muzzi, Pasupuleti, Singh, Paluri, Kardas, Tsimpoukelli, Oldham, Rita, Pavlova, Kambadur, Lewis, Si, Singh, Hassan, Goyal, Torabi, Bashlykov, Bogoychev, Chatterji, Zhang, Duchenne, Çelebi, Alrassy, Zhang, Li, Vasic, Weng, Bhargava, Dubal, Krishnan, Koura, Xu, He, Dong, Srinivasan, Ganapathy, Calderer, Cabral, Stojnic, Raileanu, Maheswari, Girdhar, Patel, Sauvestre, Polidoro, Sumbaly, Taylor, Silva, Hou, Wang, Hosseini, Chennabasappa, Singh, Bell, Kim, Edunov, Nie, Narang, Raparthy, Shen, Wan, Bhosale, Zhang, Vandenhende, Batra, Whitman, Sootla, Collot, Gururangan, Borodinsky, Herman, Fowler, Sheasha, Georgiou, Scialom, Speckbacher, Mihaylov, Xiao, Karn, Goswami, Gupta, Ramanathan, Kerkez, Gonguet, Do, Vogeti, Albiero, Petrovic, Chu, Xiong, Fu, Meers, Martinet, Wang, Wang, Tan, Xia, Xie, Jia, Wang, Goldschlag, Gaur, Babaei, Wen, Song, Zhang, Li, Mao, Coudert, Yan, Chen, Papakipos, Singh, Srivastava, Jain, Kelsey, Shajnfeld, Gangidi, Victoria, Goldstand, Menon, Sharma, Boesenberg,
  Baevski, Feinstein, Kallet, Sangani, Teo, Yunus, Lupu, Alvarado, Caples, Gu, Ho, Poulton, Ryan, Ramchandani, Dong, Franco, Goyal, Saraf, Chowdhury, Gabriel, Bharambe, Eisenman, Yazdan, James, Maurer, Leonhardi, Huang, Loyd, Paola, Paranjape, Liu, Wu, Ni, Hancock, Wasti, Spence, Stojkovic, Gamido, Montalvo, Parker, Burton, Mejia, Liu, Wang, Kim, Zhou, Hu, Chu, Cai, Tindal, Feichtenhofer, Gao, Civin, Beaty, Kreymer, Li, Adkins, Xu, Testuggine, David, Parikh, Liskovich, Foss, Wang, Le, Holland, Dowling, Jamil, Montgomery, Presani, Hahn, Wood, Le, Brinkman, Arcaute, Dunbar, Smothers, Sun, Kreuk, Tian, Kokkinos, Ozgenel, Caggioni, Kanayet, Seide, Florez, Schwarz, Badeer, Swee, Halpern, Herman, Sizov, Guangyi, Zhang, Lakshminarayanan, Inan, Shojanazeri, Zou, Wang, Zha, Habeeb, Rudolph, Suk, Aspegren, Goldman, Zhan, Damlaj, Molybog, Tufanov, Leontiadis, Veliche, Gat, Weissman, Geboski, Kohli, Lam, Asher, Gaya, Marcus, Tang, Chan, Zhen, Reizenstein, Teboul, Zhong, Jin, Yang, Cummings, Carvill, Shepard, McPhie,
  Torres, Ginsburg, Wang, Wu, U, Saxena, Khandelwal, Zand, Matosich, Veeraraghavan, Michelena, Li, Jagadeesh, Huang, Chawla, Huang, Chen, Garg, A, Silva, Bell, Zhang, Guo, Yu, Moshkovich, Wehrstedt, Khabsa, Avalani, Bhatt, Mankus, Hasson, Lennie, Reso, Groshev, Naumov, Lathi, Keneally, Liu, Seltzer, Valko, Restrepo, Patel, Vyatskov, Samvelyan, Clark, Macey, Wang, Hermoso, Metanat, Rastegari, Bansal, Santhanam, Parks, White, Bawa, Singhal, Egebo, Usunier, Mehta, Laptev, Dong, Cheng, Chernoguz, Hart, Salpekar, Kalinli, Kent, Parekh, Saab, Balaji, Rittner, Bontrager, Roux, Dollar, Zvyagina, Ratanchandani, Yuvraj, Liang, Alao, Rodriguez, Ayub, Murthy, Nayani, Mitra, Parthasarathy, Li, Hogan, Battey, Wang, Howes, Rinott, Mehta, Siby, Bondu, Datta, Chugh, Hunt, Dhillon, Sidorov, Pan, Mahajan, Verma, Yamamoto, Ramaswamy, Lindsay, Lindsay, Feng, Lin, Zha, Patil, Shankar, Zhang, Zhang, Wang, Agarwal, Sajuyigbe, Chintala, Max, Chen, Kehoe, Satterfield, Govindaprasad, Gupta, Deng, Cho, Virk, Subramanian, Choudhury,
  Goldman, Remez, Glaser, Best, Koehler, Robinson, Li, Zhang, Matthews, Chou, Shaked, Vontimitta, Ajayi, Montanez, Mohan, Kumar, Mangla, Ionescu, Poenaru, Mihailescu, Ivanov, Li, Wang, Jiang, Bouaziz, Constable, Tang, Wu, Wang, Wu, Gao, Kleinman, Chen, Hu, Jia, Qi, Li, Zhang, Zhang, Adi, Nam, Yu, Wang, Zhao, Hao, Qian, Li, He, Rait, DeVito, Rosnbrick, Wen, Yang, Zhao, and Ma}]{llama3-1}
Aaron Grattafiori, Abhimanyu Dubey, Abhinav Jauhri, Abhinav Pandey, Abhishek Kadian, Ahmad Al-Dahle, Aiesha Letman, Akhil Mathur, Alan Schelten, Alex Vaughan, Amy Yang, Angela Fan, Anirudh Goyal, Anthony Hartshorn, Aobo Yang, Archi Mitra, Archie Sravankumar, Artem Korenev, Arthur Hinsvark, and 542 others. 2024.
\newblock \href {https://arxiv.org/abs/2407.21783} {The llama 3 herd of models}.
\newblock \emph{Preprint}, arXiv:2407.21783.

\bibitem[{Guha et~al.(2025)Guha, Marten, Keh, Raoof, Smyrnis, Bansal, Nezhurina, Mercat, Vu, Sprague et~al.}]{guha2025openthoughts}
Etash Guha, Ryan Marten, Sedrick Keh, Negin Raoof, Georgios Smyrnis, Hritik Bansal, Marianna Nezhurina, Jean Mercat, Trung Vu, Zayne Sprague, and 1 others. 2025.
\newblock Openthoughts: Data recipes for reasoning models.
\newblock \emph{arXiv preprint arXiv:2506.04178}.

\bibitem[{Guo et~al.()Guo, Cao, Wang, Li, Chen, Lyu, Xu, Hu, Li et~al.}]{guosns}
Hongcheng Guo, Shaosheng Cao, Boyang Wang, Lei Li, Liang Chen, Xinze Lyu, Zhe Xu, Yao Hu, Zhoujun Li, and 1 others.
\newblock Sns-bench: Defining, building, and assessing capabilities of large language models in social networking services.
\newblock In \emph{Forty-second International Conference on Machine Learning}.

\bibitem[{Guo et~al.(2025)Guo, Zhao, Cao, Lyu, Liu, Wang, Wang, Li, Lu, Xu et~al.}]{guo2025redefining}
Hongcheng Guo, Fei Zhao, Shaosheng Cao, Xinze Lyu, Ziyan Liu, Yue Wang, Boyang Wang, Zhoujun Li, Chonggang Lu, Zhe Xu, and 1 others. 2025.
\newblock Redefining machine translation on social network services with large language models.
\newblock \emph{arXiv preprint arXiv:2504.07901}.

\bibitem[{Hendrycks et~al.(2021{\natexlab{a}})Hendrycks, Burns, Basart, Zou, Mazeika, Song, and Steinhardt}]{mmlu}
Dan Hendrycks, Collin Burns, Steven Basart, Andy Zou, Mantas Mazeika, Dawn Song, and Jacob Steinhardt. 2021{\natexlab{a}}.
\newblock Measuring massive multitask language understanding.
\newblock In \emph{{ICLR}}. OpenReview.net.

\bibitem[{Hendrycks et~al.(2021{\natexlab{b}})Hendrycks, Burns, Kadavath, Arora, Basart, Tang, Song, and Steinhardt}]{math}
Dan Hendrycks, Collin Burns, Saurav Kadavath, Akul Arora, Steven Basart, Eric Tang, Dawn Song, and Jacob Steinhardt. 2021{\natexlab{b}}.
\newblock Measuring mathematical problem solving with the {MATH} dataset.
\newblock In \emph{NeurIPS Datasets and Benchmarks}.

\bibitem[{Huang et~al.(2023)Huang, Bai, Zhu, Zhang, Zhang, Su, Liu, Lv, Zhang, Lei, Fu, Sun, and He}]{ceval}
Yuzhen Huang, Yuzhuo Bai, Zhihao Zhu, Junlei Zhang, Jinghan Zhang, Tangjun Su, Junteng Liu, Chuancheng Lv, Yikai Zhang, Jiayi Lei, Yao Fu, Maosong Sun, and Junxian He. 2023.
\newblock {C-Eval}: A multi-level multi-discipline chinese evaluation suite for foundation models.
\newblock In \emph{NeurIPS}.

\bibitem[{i~Orts(2019)}]{i2019multilingual}
{\`O}scar~Garibo i~Orts. 2019.
\newblock Multilingual detection of hate speech against immigrants and women in twitter at semeval-2019 task 5: Frequency analysis interpolation for hate in speech detection.
\newblock In \emph{Proceedings of the 13th International Workshop on Semantic Evaluation}, pages 460--463.

\bibitem[{Islam and Goldwasser(2025)}]{islam-goldwasser-2025-uncovering}
Tunazzina Islam and Dan Goldwasser. 2025.
\newblock \href {https://doi.org/10.18653/v1/2025.findings-naacl.413} {Uncovering latent arguments in social media messaging by employing {LLM}s-in-the-loop strategy}.
\newblock In \emph{Findings of the Association for Computational Linguistics: NAACL 2025}, pages 7397--7429, Albuquerque, New Mexico. Association for Computational Linguistics.

\bibitem[{Jain et~al.(2024)Jain, Han, Gu, Li, Yan, Zhang, Wang, Solar{-}Lezama, Sen, and Stoica}]{livecodebench}
Naman Jain, King Han, Alex Gu, Wen{-}Ding Li, Fanjia Yan, Tianjun Zhang, Sida Wang, Armando Solar{-}Lezama, Koushik Sen, and Ion Stoica. 2024.
\newblock {LiveCodeBench}: Holistic and contamination free evaluation of large language models for code.
\newblock \emph{CoRR}, abs/2403.07974.

\bibitem[{Jiang and Ferrara(2023)}]{jiang2023social}
Julie Jiang and Emilio Ferrara. 2023.
\newblock Social-llm: Modeling user behavior at scale using language models and social network data.
\newblock \emph{arXiv preprint arXiv:2401.00893}.

\bibitem[{Jin et~al.(2024)Jin, Choi, Verma, Wang, and Kumar}]{jin2024mm}
Yiqiao Jin, Minje Choi, Gaurav Verma, Jindong Wang, and Srijan Kumar. 2024.
\newblock Mm-soc: Benchmarking multimodal large language models in social media platforms.
\newblock \emph{arXiv preprint arXiv:2402.14154}.

\bibitem[{Kmainasi et~al.(2024)Kmainasi, Shahroor, Hasanain, Laskar, Hassan, and Alam}]{kmainasi2024llamalens}
Mohamed~Bayan Kmainasi, Ali~Ezzat Shahroor, Maram Hasanain, Sahinur~Rahman Laskar, Naeemul Hassan, and Firoj Alam. 2024.
\newblock Llamalens: Specialized multilingual llm for analyzing news and social media content.
\newblock \emph{arXiv preprint arXiv:2410.15308}.

\bibitem[{Konstantinidis et~al.(2024)Konstantinidis, Iacovides, Xu, Constantinides, and Mandic}]{konstantinidis2024finllama}
Thanos Konstantinidis, Giorgos Iacovides, Mingxue Xu, Tony~G Constantinides, and Danilo Mandic. 2024.
\newblock Finllama: Financial sentiment classification for algorithmic trading applications.
\newblock \emph{arXiv preprint arXiv:2403.12285}.

\bibitem[{Kumar et~al.(2024)Kumar, AbuHashem, and Durumeric}]{kumar2024watch}
Deepak Kumar, Yousef~Anees AbuHashem, and Zakir Durumeric. 2024.
\newblock Watch your language: Investigating content moderation with large language models.
\newblock In \emph{Proceedings of the International AAAI Conference on Web and Social Media}, volume~18, pages 865--878.

\bibitem[{Li et~al.(2023{\natexlab{a}})Li, Zhang, Koto, Yang, Zhao, Gong, Duan, and Baldwin}]{cmmlu}
Haonan Li, Yixuan Zhang, Fajri Koto, Yifei Yang, Hai Zhao, Yeyun Gong, Nan Duan, and Timothy Baldwin. 2023{\natexlab{a}}.
\newblock {CMMLU}: Measuring massive multitask language understanding in {Chinese}.
\newblock \emph{CoRR}, abs/2306.09212.

\bibitem[{Li et~al.(2025)Li, Du, Zhao, Zhang, Wang, Gao, Liu, and Lin}]{li2025infinity}
Jijie Li, Li~Du, Hanyu Zhao, Bo-wen Zhang, Liangdong Wang, Boyan Gao, Guang Liu, and Yonghua Lin. 2025.
\newblock Infinity instruct: Scaling instruction selection and synthesis to enhance language models.
\newblock \emph{arXiv preprint arXiv:2506.11116}.

\bibitem[{Li et~al.(2023{\natexlab{b}})Li, Cheng, Zhao, Nie, and Wen}]{halueval}
Junyi Li, Xiaoxue Cheng, Xin Zhao, Jian-Yun Nie, and Ji-Rong Wen. 2023{\natexlab{b}}.
\newblock \href {https://doi.org/10.18653/v1/2023.emnlp-main.397} {{H}alu{E}val: A large-scale hallucination evaluation benchmark for large language models}.
\newblock pages 6449--6464.

\bibitem[{Li et~al.(2024{\natexlab{a}})Li, Chen, Tang, ShengbinHou, Wang, Deng, Li, Xiong, Mao, Peng, and Luo}]{newsbench}
Miao Li, Ming-Bin Chen, Bo~Tang, ShengbinHou ShengbinHou, Pengyu Wang, Haiying Deng, Zhiyu Li, Feiyu Xiong, Keming Mao, Cheng Peng, and Yi~Luo. 2024{\natexlab{a}}.
\newblock \href {https://doi.org/10.18653/v1/2024.acl-long.538} {{N}ews{B}ench: A systematic evaluation framework for assessing editorial capabilities of large language models in {C}hinese journalism}.
\newblock pages 9993--10014.

\bibitem[{Li et~al.(2024{\natexlab{b}})Li, Meng, Zhou, Wei, Gan, and Pfister}]{li2024socialgpt}
Wanhua Li, Zibin Meng, Jiawei Zhou, Donglai Wei, Chuang Gan, and Hanspeter Pfister. 2024{\natexlab{b}}.
\newblock Socialgpt: Prompting llms for social relation reasoning via greedy segment optimization.
\newblock \emph{arXiv preprint arXiv:2410.21411}.

\bibitem[{Lin et~al.(2024)Lin, Luo, Gao, Ma, Wang, and Yang}]{lin2024towards}
Hongzhan Lin, Ziyang Luo, Wei Gao, Jing Ma, Bo~Wang, and Ruichao Yang. 2024.
\newblock Towards explainable harmful meme detection through multimodal debate between large language models.
\newblock In \emph{Proceedings of the ACM Web Conference 2024}, pages 2359--2370.

\bibitem[{Liu et~al.(2024{\natexlab{a}})Liu, Zheng, Muennighoff, Zeng, Dou, Pang, Jiang, and Lin}]{liu2024regmix}
Qian Liu, Xiaosen Zheng, Niklas Muennighoff, Guangtao Zeng, Longxu Dou, Tianyu Pang, Jing Jiang, and Min Lin. 2024{\natexlab{a}}.
\newblock Regmix: Data mixture as regression for language model pre-training.
\newblock \emph{arXiv preprint arXiv:2407.01492}.

\bibitem[{Liu et~al.(2024{\natexlab{b}})Liu, Tao, Wu, Wu, and Wang}]{liu2024can}
Qiang Liu, Xiang Tao, Junfei Wu, Shu Wu, and Liang Wang. 2024{\natexlab{b}}.
\newblock Can large language models detect rumors on social media?
\newblock \emph{arXiv preprint arXiv:2402.03916}.

\bibitem[{Loshchilov and Hutter(2017)}]{adamw}
Ilya Loshchilov and Frank Hutter. 2017.
\newblock Decoupled weight decay regularization.
\newblock \emph{CoRR}, abs/1711.05101.

\bibitem[{Lu et~al.(2024)Lu, Xu, Zhang, Wang, Zhu, Zhang, Yang, and Lin}]{lu2024towards}
Junyu Lu, Bo~Xu, Xiaokun Zhang, Hongbo Wang, Haohao Zhu, Dongyu Zhang, Liang Yang, and Hongfei Lin. 2024.
\newblock Towards comprehensive detection of chinese harmful memes.
\newblock \emph{Advances in Neural Information Processing Systems}, 37:13302--13320.

\bibitem[{{MAA}(2025)}]{aime25}
{MAA}. 2025.
\newblock \href {https://maa.org/math-competitions/american-invitational-mathematics-examination-aime} {American invitational mathematics examination - aime}.
\newblock \emph{American Invitational Mathematics Examination - AIME 2025}.

\bibitem[{McCloskey and Cohen(1989)}]{mccloskey1989catastrophic}
Michael McCloskey and Neal~J Cohen. 1989.
\newblock Catastrophic interference in connectionist networks: The sequential learning problem.
\newblock In \emph{Psychology of learning and motivation}, volume~24, pages 109--165. Elsevier.

\bibitem[{Mistral-AI(2024)}]{ministral}
Mistral-AI. 2024.
\newblock Un ministral, des ministraux.
\newblock \url{https://mistral.ai/news/ministraux}.
\newblock Accessed: 2024-10-16.

\bibitem[{Mistral-AI(2025)}]{mistralsmall2025}
Mistral-AI. 2025.
\newblock Mistral small 3.1.
\newblock \url{https://mistral.ai/news/mistral-small-3-1}.
\newblock Accessed: 2025-03-17.

\bibitem[{Mohammad et~al.(2018)Mohammad, Bravo-Marquez, Salameh, and Kiritchenko}]{mohammad2018semeval}
Saif Mohammad, Felipe Bravo-Marquez, Mohammad Salameh, and Svetlana Kiritchenko. 2018.
\newblock Semeval-2018 task 1: Affect in tweets.
\newblock In \emph{Proceedings of the 12th international workshop on semantic evaluation}, pages 1--17.

\bibitem[{OpenAI()}]{gpt4o-1120}
OpenAI.
\newblock Gpt-4o: Openai’s newest multimodal model.
\newblock \url{https://openai.com/index/gpt-4o}.

\bibitem[{Ouyang et~al.(2022)Ouyang, Wu, Jiang, Almeida, Wainwright, Mishkin, Zhang, Agarwal, Slama, Ray et~al.}]{ouyang2022training}
Long Ouyang, Jeffrey Wu, Xu~Jiang, Diogo Almeida, Carroll Wainwright, Pamela Mishkin, Chong Zhang, Sandhini Agarwal, Katarina Slama, Alex Ray, and 1 others. 2022.
\newblock Training language models to follow instructions with human feedback.
\newblock \emph{Advances in neural information processing systems}, 35:27730--27744.

\bibitem[{Penedo et~al.(2024)Penedo, Kydl{\'\i}{\v{c}}ek, Lozhkov, Mitchell, Raffel, Von~Werra, Wolf et~al.}]{penedo2024fineweb}
Guilherme Penedo, Hynek Kydl{\'\i}{\v{c}}ek, Anton Lozhkov, Margaret Mitchell, Colin~A Raffel, Leandro Von~Werra, Thomas Wolf, and 1 others. 2024.
\newblock The fineweb datasets: Decanting the web for the finest text data at scale.
\newblock \emph{Advances in Neural Information Processing Systems}, 37:30811--30849.

\bibitem[{Peng et~al.(2024)Peng, Wang, Yao, Wang, and Shang}]{peng2024metaie}
Letian Peng, Zilong Wang, Feng Yao, Zihan Wang, and Jingbo Shang. 2024.
\newblock Metaie: Distilling a meta model from llm for all kinds of information extraction tasks.
\newblock \emph{arXiv preprint arXiv:2404.00457}.

\bibitem[{Qiu et~al.(2024)Qiu, Lv, Jin, Wang, Ning, Yu, Zhang, Li, Chu, Qu et~al.}]{qiu2024wanjuan}
Jiantao Qiu, Haijun Lv, Zhenjiang Jin, Rui Wang, Wenchang Ning, Jia Yu, ChaoBin Zhang, Zhenxiang Li, Pei Chu, Yuan Qu, and 1 others. 2024.
\newblock Wanjuan-cc: A safe and high-quality open-sourced english webtext dataset.
\newblock \emph{arXiv preprint arXiv:2402.19282}.

\bibitem[{Qwen et~al.(2025{\natexlab{a}})Qwen, :, Yang, Yang, Zhang, Hui, Zheng, Yu, Li, Liu, Huang, Wei, Lin, Yang, Tu, Zhang, Yang, Yang, Zhou, Lin, Dang, Lu, Bao, Yang, Yu, Li, Xue, Zhang, Zhu, Men, Lin, Li, Tang, Xia, Ren, Ren, Fan, Su, Zhang, Wan, Liu, Cui, Zhang, and Qiu}]{qwen2025qwen25technicalreport}
Qwen, :, An~Yang, Baosong Yang, Beichen Zhang, Binyuan Hui, Bo~Zheng, Bowen Yu, Chengyuan Li, Dayiheng Liu, Fei Huang, Haoran Wei, Huan Lin, Jian Yang, Jianhong Tu, Jianwei Zhang, Jianxin Yang, Jiaxi Yang, Jingren Zhou, and 25 others. 2025{\natexlab{a}}.
\newblock \href {https://arxiv.org/abs/2412.15115} {Qwen2.5 technical report}.
\newblock \emph{Preprint}, arXiv:2412.15115.

\bibitem[{Qwen et~al.(2025{\natexlab{b}})Qwen, :, Yang, Yang, Zhang, Hui, Zheng, Yu, Li, Liu, Huang, Wei, Lin, Yang, Tu, Zhang, Yang, Yang, Zhou, Lin, Dang, Lu, Bao, Yang, Yu, Li, Xue, Zhang, Zhu, Men, Lin, Li, Tang, Xia, Ren, Ren, Fan, Su, Zhang, Wan, Liu, Cui, Zhang, and Qiu}]{qwen2-5}
Qwen, :, An~Yang, Baosong Yang, Beichen Zhang, Binyuan Hui, Bo~Zheng, Bowen Yu, Chengyuan Li, Dayiheng Liu, Fei Huang, Haoran Wei, Huan Lin, Jian Yang, Jianhong Tu, Jianwei Zhang, Jianxin Yang, Jiaxi Yang, Jingren Zhou, and 25 others. 2025{\natexlab{b}}.
\newblock \href {https://arxiv.org/abs/2412.15115} {Qwen2.5 technical report}.

\bibitem[{Rafailov et~al.(2023)Rafailov, Sharma, Mitchell, Manning, Ermon, and Finn}]{rafailov2024direct}
Rafael Rafailov, Archit Sharma, Eric Mitchell, Christopher~D. Manning, Stefano Ermon, and Chelsea Finn. 2023.
\newblock Direct preference optimization: Your language model is secretly a reward model.
\newblock In \emph{NeurIPS}.

\bibitem[{Rein et~al.(2023)Rein, Hou, Stickland, Petty, Pang, Dirani, Michael, and Bowman}]{gpqa}
David Rein, Betty~Li Hou, Asa~Cooper Stickland, Jackson Petty, Richard~Yuanzhe Pang, Julien Dirani, Julian Michael, and Samuel~R. Bowman. 2023.
\newblock {GPQA}: A graduate-level {Google}-proof {Q}{\&}{A} benchmark.
\newblock \emph{CoRR}, abs/2311.12022.

\bibitem[{Rosenthal et~al.(2019)Rosenthal, Farra, and Nakov}]{rosenthal2019semeval}
Sara Rosenthal, Noura Farra, and Preslav Nakov. 2019.
\newblock Semeval-2017 task 4: Sentiment analysis in twitter.
\newblock \emph{arXiv preprint arXiv:1912.00741}.

\bibitem[{Suzgun et~al.(2023)Suzgun, Scales, Sch{\"{a}}rli, Gehrmann, Tay, Chung, Chowdhery, Le, Chi, Zhou, and Wei}]{bbh}
Mirac Suzgun, Nathan Scales, Nathanael Sch{\"{a}}rli, Sebastian Gehrmann, Yi~Tay, Hyung~Won Chung, Aakanksha Chowdhery, Quoc~V. Le, Ed~H. Chi, Denny Zhou, and Jason Wei. 2023.
\newblock Challenging {BIG-Bench} tasks and whether chain-of-thought can solve them.
\newblock In \emph{{ACL} (Findings)}, pages 13003--13051. Association for Computational Linguistics.

\bibitem[{Wang et~al.(2024)Wang, Ma, Zhang, Ni, Chandra, Guo, Ren, Arulraj, He, Jiang, Li, Ku, Wang, Zhuang, Fan, Yue, and Chen}]{mmlupro}
Yubo Wang, Xueguang Ma, Ge~Zhang, Yuansheng Ni, Abhranil Chandra, Shiguang Guo, Weiming Ren, Aaran Arulraj, Xuan He, Ziyan Jiang, Tianle Li, Max Ku, Kai Wang, Alex Zhuang, Rongqi Fan, Xiang Yue, and Wenhu Chen. 2024.
\newblock {MMLU-Pro}: {A} more robust and challenging multi-task language understanding benchmark.
\newblock \emph{CoRR}, abs/2406.01574.

\bibitem[{Wang et~al.(2025)Wang, Fu, Cai, Tang, Lyu, Fang, Zheng, Zhou, Zeng, Xiao et~al.}]{wang2025ultra}
Yudong Wang, Zixuan Fu, Jie Cai, Peijun Tang, Hongya Lyu, Yewei Fang, Zhi Zheng, Jie Zhou, Guoyang Zeng, Chaojun Xiao, and 1 others. 2025.
\newblock Ultra-fineweb: Efficient data filtering and verification for high-quality llm training data.
\newblock \emph{arXiv preprint arXiv:2505.05427}.

\bibitem[{Weber et~al.(2024)Weber, Fu, Anthony, Oren, Adams, Alexandrov, Lyu, Nguyen, Yao, Adams et~al.}]{weber2024redpajama}
Maurice Weber, Dan Fu, Quentin Anthony, Yonatan Oren, Shane Adams, Anton Alexandrov, Xiaozhong Lyu, Huu Nguyen, Xiaozhe Yao, Virginia Adams, and 1 others. 2024.
\newblock Redpajama: an open dataset for training large language models.
\newblock \emph{Advances in neural information processing systems}, 37:116462--116492.

\bibitem[{Wen et~al.(2023)Wen, Sun, Zhao, Fang, Chen, and Zou}]{wen2023chathome}
Cheng Wen, Xianghui Sun, Shuaijiang Zhao, Xiaoquan Fang, Liangyu Chen, and Wei Zou. 2023.
\newblock Chathome: Development and evaluation of a domain-specific language model for home renovation.
\newblock \emph{arXiv preprint arXiv:2307.15290}.

\bibitem[{Wu et~al.(2024)Wu, Lin, Zhang, Zhang, Xie, and Wang}]{wu2024pmc}
Chaoyi Wu, Weixiong Lin, Xiaoman Zhang, Ya~Zhang, Weidi Xie, and Yanfeng Wang. 2024.
\newblock Pmc-llama: toward building open-source language models for medicine.
\newblock \emph{Journal of the American Medical Informatics Association}, 31(9):1833--1843.

\bibitem[{Wu et~al.(2023)Wu, Irsoy, Lu, Dabravolski, Dredze, Gehrmann, Kambadur, Rosenberg, and Mann}]{wu2023bloomberggpt}
Shijie Wu, Ozan Irsoy, Steven Lu, Vadim Dabravolski, Mark Dredze, Sebastian Gehrmann, Prabhanjan Kambadur, David Rosenberg, and Gideon Mann. 2023.
\newblock Bloomberggpt: A large language model for finance.
\newblock \emph{arXiv preprint arXiv:2303.17564}.

\bibitem[{Xie et~al.(2023)Xie, He, Xu, Wu, Zhu, Yang, and Chen}]{10.1145/3612920}
Zheyong Xie, Weidong He, Tong Xu, Shiwei Wu, Chen Zhu, Ping Yang, and Enhong Chen. 2023.
\newblock \href {https://doi.org/10.1145/3612920} {Comprehending the gossips: Meme explanation in time-sync video comment via multimodal cues}.
\newblock \emph{ACM Trans. Asian Low-Resour. Lang. Inf. Process.}, 22(8).

\bibitem[{Xiong et~al.(2023)Xiong, Wang, Zhu, Zhao, Liu, Huang, Wang, and Shen}]{xiong2023doctorglm}
Honglin Xiong, Sheng Wang, Yitao Zhu, Zihao Zhao, Yuxiao Liu, Linlin Huang, Qian Wang, and Dinggang Shen. 2023.
\newblock Doctorglm: Fine-tuning your chinese doctor is not a herculean task.
\newblock \emph{arXiv preprint arXiv:2304.01097}.

\bibitem[{Yang et~al.(2024{\natexlab{a}})Yang, Yang, Hui, Zheng, Yu, Zhou, Li, Li, Liu, Huang, Dong, Wei, Lin, Tang, Wang, Yang, Tu, Zhang, Ma, Yang, Xu, Zhou, Bai, He, Lin, Dang, Lu, Chen, Yang, Li, Xue, Ni, Zhang, Wang, Peng, Men, Gao, Lin, Wang, Bai, Tan, Zhu, Li, Liu, Ge, Deng, Zhou, Ren, Zhang, Wei, Ren, Liu, Fan, Yao, Zhang, Wan, Chu, Liu, Cui, Zhang, Guo, and Fan}]{qwen2}
An~Yang, Baosong Yang, Binyuan Hui, Bo~Zheng, Bowen Yu, Chang Zhou, Chengpeng Li, Chengyuan Li, Dayiheng Liu, Fei Huang, Guanting Dong, Haoran Wei, Huan Lin, Jialong Tang, Jialin Wang, Jian Yang, Jianhong Tu, Jianwei Zhang, Jianxin Ma, and 43 others. 2024{\natexlab{a}}.
\newblock Qwen2 technical report.
\newblock \emph{CoRR}, abs/2407.10671.

\bibitem[{Yang et~al.(2024{\natexlab{b}})Yang, Zhang, Kuang, Xie, Huang, and Ananiadou}]{yang2024mentallama}
Kailai Yang, Tianlin Zhang, Ziyan Kuang, Qianqian Xie, Jimin Huang, and Sophia Ananiadou. 2024{\natexlab{b}}.
\newblock Mentallama: interpretable mental health analysis on social media with large language models.
\newblock In \emph{Proceedings of the ACM Web Conference 2024}, pages 4489--4500.

\bibitem[{Yang et~al.(2024{\natexlab{c}})Yang, Zhao, Zhu, Zhou, Xu, Jia, and Zan}]{yang2024zhongjing}
Songhua Yang, Hanjie Zhao, Senbin Zhu, Guangyu Zhou, Hongfei Xu, Yuxiang Jia, and Hongying Zan. 2024{\natexlab{c}}.
\newblock Zhongjing: Enhancing the chinese medical capabilities of large language model through expert feedback and real-world multi-turn dialogue.
\newblock In \emph{Proceedings of the AAAI conference on artificial intelligence}, volume~38, pages 19368--19376.

\bibitem[{Yang et~al.(2024{\natexlab{d}})Yang, Gao, Xue, and Alexandersson}]{yang2024pllama}
Xianjun Yang, Junfeng Gao, Wenxin Xue, and Erik Alexandersson. 2024{\natexlab{d}}.
\newblock Pllama: An open-source large language model for plant science.
\newblock \emph{arXiv preprint arXiv:2401.01600}.

\bibitem[{Ye et~al.(2025)Ye, Huang, Xiao, Chern, Xia, and Liu}]{ye2025limo}
Yixin Ye, Zhen Huang, Yang Xiao, Ethan Chern, Shijie Xia, and Pengfei Liu. 2025.
\newblock Limo: Less is more for reasoning.
\newblock \emph{arXiv preprint arXiv:2502.03387}.

\bibitem[{Yi et~al.(2024)Yi, Ouyang, Liu, Liao, Xu, and Shen}]{yi2024survey}
Zihao Yi, Jiarui Ouyang, Yuwen Liu, Tianhao Liao, Zhe Xu, and Ying Shen. 2024.
\newblock A survey on recent advances in llm-based multi-turn dialogue systems.
\newblock \emph{arXiv preprint arXiv:2402.18013}.

\bibitem[{Yuan et~al.(2024)Yuan, Rastogi, Naik, Rajagopal, Goyal, Zhao, Chintagunta, and Ward}]{yuan2024continued}
Dong Yuan, Eti Rastogi, Gautam Naik, Sree~Prasanna Rajagopal, Sagar Goyal, Fen Zhao, Bharath Chintagunta, and Jeff Ward. 2024.
\newblock A continued pretrained llm approach for automatic medical note generation.
\newblock In \emph{NAACL (Short Papers)}.

\bibitem[{Yue et~al.(2025)Yue, Chen, Lu, Zhao, Wang, Yue, Song, and Huang}]{yue2025does}
Yang Yue, Zhiqi Chen, Rui Lu, Andrew Zhao, Zhaokai Wang, Yang Yue, Shiji Song, and Gao Huang. 2025.
\newblock Does reinforcement learning really incentivize reasoning capacity in llms beyond the base model?
\newblock \emph{arXiv e-prints}, pages arXiv--2504.

\bibitem[{Zakka et~al.(2024)Zakka, Shad, Chaurasia, Dalal, Kim, Moor, Fong, Phillips, Alexander, Ashley et~al.}]{zakka2024almanac}
Cyril Zakka, Rohan Shad, Akash Chaurasia, Alex~R Dalal, Jennifer~L Kim, Michael Moor, Robyn Fong, Curran Phillips, Kevin Alexander, Euan Ashley, and 1 others. 2024.
\newblock Almanac—retrieval-augmented language models for clinical medicine.
\newblock \emph{Nejm ai}, 1(2):AIoa2300068.

\bibitem[{Zeng et~al.(2024)Zeng, Huang, Malik, Yin, Babic, Shacham, Yan, Yang, and He}]{zeng2024large}
Jingying Zeng, Richard Huang, Waleed Malik, Langxuan Yin, Bojan Babic, Danny Shacham, Xiao Yan, Jaewon Yang, and Qi~He. 2024.
\newblock Large language models for social networks: Applications, challenges, and solutions.
\newblock \emph{arXiv preprint arXiv:2401.02575}.

\bibitem[{Zhang et~al.(2024)Zhang, Naradowsky, and Miyao}]{zhang-etal-2024-self-emotion}
Qiang Zhang, Jason Naradowsky, and Yusuke Miyao. 2024.
\newblock \href {https://doi.org/10.18653/v1/2024.sigdial-1.21} {Self-emotion blended dialogue generation in social simulation agents}.
\newblock In \emph{Proceedings of the 25th Annual Meeting of the Special Interest Group on Discourse and Dialogue}, pages 228--247, Kyoto, Japan. Association for Computational Linguistics.

\bibitem[{Zhang et~al.(2023)Zhang, Li, Zong, Ying, He, and Qiu}]{gaokao-bench}
Xiaotian Zhang, Chunyang Li, Yi~Zong, Zhengyu Ying, Liang He, and Xipeng Qiu. 2023.
\newblock \href {https://doi.org/10.48550/arXiv.2305.12474} {Evaluating the performance of large language models on {GAOKAO} benchmark}.
\newblock \emph{CoRR}, abs/2305.12474.

\bibitem[{Zhao et~al.(2024)Zhao, Ren, Hessel, Cardie, Choi, and Deng}]{zhao2024wildchat}
Wenting Zhao, Xiang Ren, Jack Hessel, Claire Cardie, Yejin Choi, and Yuntian Deng. 2024.
\newblock Wildchat: 1m chatgpt interaction logs in the wild.
\newblock \emph{arXiv preprint arXiv:2405.01470}.

\bibitem[{Zheng et~al.(2024)Zheng, Zhang, Zhang, Ye, Luo, Feng, and Ma}]{zheng2024llamafactory}
Yaowei Zheng, Richong Zhang, Junhao Zhang, Yanhan Ye, Zheyan Luo, Zhangchi Feng, and Yongqiang Ma. 2024.
\newblock Llamafactory: Unified efficient fine-tuning of 100+ language models.
\newblock \emph{arXiv preprint arXiv:2403.13372}.

\bibitem[{Zhou et~al.(2023{\natexlab{a}})Zhou, Liu, Xu, Iyer, Sun, Mao, Ma, Efrat, Yu, Yu et~al.}]{zhou2023lima}
Chunting Zhou, Pengfei Liu, Puxin Xu, Srinivasan Iyer, Jiao Sun, Yuning Mao, Xuezhe Ma, Avia Efrat, Ping Yu, Lili Yu, and 1 others. 2023{\natexlab{a}}.
\newblock Lima: Less is more for alignment.
\newblock \emph{Advances in Neural Information Processing Systems}, 36:55006--55021.

\bibitem[{Zhou et~al.(2023{\natexlab{b}})Zhou, Lu, Mishra, Brahma, Basu, Luan, Zhou, and Hou}]{ifeval}
Jeffrey Zhou, Tianjian Lu, Swaroop Mishra, Siddhartha Brahma, Sujoy Basu, Yi~Luan, Denny Zhou, and Le~Hou. 2023{\natexlab{b}}.
\newblock Instruction-following evaluation for large language models.
\newblock \emph{CoRR}, abs/2311.07911.

\end{thebibliography}
